\documentclass{article}

\usepackage{microtype}
\usepackage{graphicx}
\usepackage{booktabs}
\usepackage{multirow}
\usepackage{hyperref}
\usepackage{tabularx}
\usepackage{array}
\usepackage{xcolor}
\usepackage{colortbl}
\usepackage{tikz}
\usepackage{float}
\usetikzlibrary{arrows.meta,positioning,calc,fit}

\usepackage[accepted]{icml2026}

\makeatletter
\renewcommand{\ICML@appearing}{\textit{3rd AI for Math Workshop at the 43${}^{rd}$ International Conference on Machine Learning}, Seoul, South Korea, 2026.}
\makeatother

\makeatletter
\let\ICML@appearing@orig\ICML@appearing
\renewcommand{\ICML@appearing}{%
  \ICML@appearing@orig\\
  {\raggedright Code and data: \url{https://github.com/vatsalananthula/verifiedcoupling}\par}%
}
\makeatother

\usepackage{amsmath}
\usepackage{amssymb}
\usepackage{mathtools}
\usepackage{amsthm}
\usepackage[capitalize,noabbrev]{cleveref}

\usepackage{titlesec}
\titlespacing*{\section}{0pt}{5pt plus 1pt minus 1pt}{2pt plus 1pt minus 1pt}
\titlespacing*{\subsection}{0pt}{4pt plus 1pt minus 1pt}{1pt plus 1pt minus 1pt}

\definecolor{vcTeal}{HTML}{20808D}
\definecolor{vcDarkTeal}{HTML}{1B474D}
\definecolor{vcRust}{HTML}{A84B2F}
\definecolor{vcGold}{HTML}{D19900}
\definecolor{vcGray}{HTML}{D4D1CA}
\definecolor{vcText}{HTML}{28251D}
\definecolor{vcSoft}{HTML}{EEF6F7}
\definecolor{vcWarm}{HTML}{FFF8E5}

\icmltitlerunning{Decodable but Not Faithful: Coupling Rationales to Programmatic Verifiers}

\begin{document}

\twocolumn[
\icmltitle{Decodable but Not Faithful: Coupling Natural-Language Rationales to Programmatic Verifiers}

\begin{icmlauthorlist}
  \icmlauthor{Vatsal Ananthula}{ucb}
  \icmlauthor{Adarsh Kumarappan}{caltech}
\end{icmlauthorlist}

\icmlaffiliation{ucb}{University of California, Berkeley}
\icmlaffiliation{caltech}{California Institute of Technology}
\icmlcorrespondingauthor{Vatsal Ananthula}{vananthu@berkeley.edu}
\icmlcorrespondingauthor{Adarsh Kumarappan}{adarsh@caltech.edu}
\icmlkeywords{AI4Math, verification, reasoning, process supervision, programmatic verifiers, reinforcement learning}

\vskip 0.2in
]

\printAffiliationsAndNotice{}

\begin{abstract}
Language models can generate plausible rationales for their predictions, but these explanations may not faithfully represent the model's internal reasoning. We propose \emph{verifier-coupled reasoning}, a framework that inserts inline claims into reasoning traces and trains an auxiliary consistency head to predict programmatic verifier outputs from rationale-span hidden states. The central finding is a gap between \emph{decodability} and \emph{faithfulness}: consistency training reliably makes verifier information decodable from rationale representations, but decodability does not guarantee faithful generation. In LeanCheck (formal theorem proving), rationale-only and proof-only pooling achieve perfect directional separation under counterfactual conflict. In KataGo (Go engine), commentary spans encode 10-way win-rate buckets at 81\% accuracy. Yet in a code setting, the model achieves 98.6\% coupling while its generated explanations remain unfaithful: fluent prose with correct structured claims, but describing unrelated algorithms; a controlled pretrained-vs-from-scratch comparison shows the gap is not capacity-driven. Synthetic activation patching confirms causal influence (73--89\% vs.\ 31\% baseline), FEVER reveals that evidence-only pooling isolates genuine evidence sensitivity at the cost of raw accuracy, and per-claim analysis shows that consistency loss disproportionately benefits fine-grained claims over binary ones. These results establish that consistency losses are effective diagnostics and representation-shaping tools, but not sufficient conditions for faithful reasoning.
\end{abstract}

\section{Introduction}

Chain-of-thought and rationale generation have become default tools for making language-model behavior more inspectable, especially on mathematical and scientific reasoning tasks. The appeal is straightforward: if a model exposes intermediate reasoning, a user or verifier can inspect the path rather than only the final answer. Prior work has shown that chain-of-thought can improve task accuracy \citep{wei2022cot}, and self-consistency can further improve reasoning by aggregating sampled reasoning paths \citep{wang2022selfconsistency}. Yet this transparency is not perfect. A model can write a plausible explanation that rationalizes a decision after the fact, omits the true source of an answer, or uses shortcuts invisible in the generated text. Generated explanations may fail to faithfully represent the model's internal decision process \citep{turpin2023unfaithful}, and probing work cautions that decodable information is not automatically evidence of causal use \citep{hewitt2019designing}.

External verifiers offer an opportunity to address these issues. In mathematical reasoning, theorem provers and symbolic systems can check claims. In factual verification, evidence labels can serve as targets. In games and scientific domains, engines or simulators can compute dense state evaluations. Process supervision shows that step-level feedback can improve mathematical reasoning beyond outcome-only supervision \citep{lightman2023letsverify}, and recent RL with verifiable rewards uses automatically checkable outcomes to train stronger reasoning policies \citep{deepseek2025r1}. The usual use of these tools is outcome-level: check the final answer, rerank candidates, or train a reward model. That leaves a gap between \emph{verified claims} and \emph{natural-language rationales}. The model may produce a correct claim while the explanation is unsupported, or encode verifier information internally while failing to verbalize it coherently.

This paper studies a mechanism for narrowing this gap by making the rationale representation directly accountable to the verifier. Rather than only checking the final claim, we train the hidden states of the rationale span to predict the verifier's output. This creates a measurable intermediate condition: \emph{decodability}, meaning that verifier information can be decoded from rationale states. Decodability is a necessary condition for faithfulness---a rationale whose hidden states carry no verifier-relevant information cannot faithfully explain the claim---but the critical question is whether it is sufficient. Does a model whose representations encode verifier information also \emph{generate} explanations that faithfully reflect it?

Concretely, the model emits a sequence containing a natural-language rationale and an explicit inline claim. A programmatic verifier supplies a target label or scalar. An auxiliary consistency head---a linear classifier over mean-pooled rationale hidden states---predicts that target. The resulting consistency loss is added to the standard language-modeling objective. We then escalate diagnostics from decodability (consistency-head accuracy) through counterfactual following (do swapped spans change predictions?) to causal influence (does patching rationale activations shift claim logits?) and generation quality (does the model produce usable explanations?).

The central finding is that decodability and faithfulness come apart. In controlled and formal-verifier settings, consistency training makes verifier information reliably decodable and even causally influential under activation patching. But in a code-explanation setting, the model achieves 98.6\% representation coupling while its generated prose, though fluent, describes unrelated algorithms; a controlled comparison on a pretrained GPT-2 reproduces the same fluent-but-unfaithful failure mode, showing the gap is structural rather than a capacity artifact. This separation of representation and generation is the paper's main result, and it has implications for any approach that uses probe accuracy as evidence of faithful reasoning.

We test across five settings: synthetic (controlled validation and causal intervention), LeanCheck (Lean theorem prover), KataGo (Go engine), code coupling (where the failure mode is most visible), and FEVER (real-language evidence sensitivity). Our contributions are:
\begin{itemize}\setlength{\itemsep}{1pt}\setlength{\parskip}{0pt}\setlength{\topsep}{2pt}
    \item \textbf{Verifier-coupled reasoning}: a sequence format and span-level consistency loss for coupling natural-language rationales to programmatic verifier claims, applicable to theorem provers, game engines, code oracles, and evidence labels.
    \item \textbf{A diagnostic ladder from decodability to faithfulness}: consistency-head accuracy, counterfactual span swaps, activation patching, per-claim analysis, and generation-quality evaluation, each level revealing information the previous level misses.
    \item \textbf{A characterization of the decodability--faithfulness gap}: across five settings, representation coupling reaches near-perfect levels while generated explanations can remain unusable, establishing that SFT-style consistency losses are effective diagnostics but not sufficient for faithful reasoning.
\end{itemize}

\section{Related Work}

\textbf{Rationales and chain-of-thought faithfulness.}
Chain-of-thought prompting demonstrates that intermediate reasoning can improve language-model performance on arithmetic, symbolic, and commonsense tasks \citep{wei2022cot}, and self-consistency improves reasoning by marginalizing over sampled reasoning paths \citep{wang2022selfconsistency}. However, generated rationales need not reveal the causal basis of a model's answer. Models can rationalize biased answers without mentioning the biasing feature \citep{turpin2023unfaithful}, and faithfulness can remain limited even after in-context learning, fine-tuning, or activation editing \citep{tanneru2024faithfulcot}. Broader analyses of explanation faithfulness emphasize that plausible natural language should not be equated with faithful computation \citep{jacovi2020faithful,lanham2023measuring}.

\textbf{Process supervision and verifier-guided generation.}
Process supervision trains systems to evaluate or supervise intermediate reasoning steps rather than only final answers. Step-level supervision can outperform outcome supervision on mathematical reasoning benchmarks \citep{lightman2023letsverify}, and automated process-supervision methods reduce dependence on human labels through search, rollout, or code verification \citep{uesato2022solving,cobbe2021training,wang2023mathshepherd,luo2024omegaprm}. Verifier-guided reasoning also appears in code and scientific settings, where execution or simulation can provide feedback \citep{chen2021codex,dong2024selfplay}. Our work differs in where the pressure is applied: rather than only scoring completed steps, we train the hidden representation of the natural-language rationale span itself to encode the verifier-derived claim. Lean 4 provides a natural formal-verifier target because it combines an interactive theorem prover with an extensible programming language \citep{demoura2021lean4}. Recent work has also demonstrated the effectiveness of LLMs integrated with Lean for lifelong formal theorem proving across diverse mathematical domains~\citep{kumarappanLeanAgent2024}.

\textbf{RLHF, RLVR, and verifiable rewards.}
RLHF aligns models to human preferences by training reward models and optimizing policies \citep{christiano2017deep,ouyang2022training}. More recent reasoning systems increasingly use verifiable rewards from math, code, or structured environments \citep{deepseek2025r1,shao2024deepseekmath,lambert2024t}. Verifier-coupled reasoning can be viewed as a supervised precursor to RL with verifiable rewards: it creates inline claims that can later be checked and rewarded automatically.

\textbf{Probing, causal interventions, and activation patching.}
Our work is most closely related to probing and causal representation evaluation. Probing warns that high probe accuracy may reflect probe capacity rather than genuine encoded structure, motivating controls and selectivity metrics \citep{hewitt2019designing,pimentel2020information}. Causal tracing and activation patching provide stronger tests by asking whether replacing or ablating representations changes model outputs \citep{meng2022locating,zhang2023activation}. Truth-probing work shows that factual properties can sometimes be linearly represented and manipulated in activations \citep{marks2023geometry}. Our work differs from standard probing in two ways. First, we do not merely \emph{decode} information from pre-existing representations; we \emph{train} the representation to encode verifier-derived targets via the consistency loss, then test whether this trained coupling transfers to generation behavior. Second, we systematically escalate diagnostics: from decodability (consistency-head accuracy) through counterfactual following (do swapped spans change the head's prediction?) to causal influence (does patching rationale activations shift claim logits?) to generation quality (does the model produce coherent explanations?). The code experiment demonstrates why this escalation matters: decodability reaches 98.6\% while generation quality is zero, a gap that standard probing alone would not reveal.

\section{Background}

\textbf{Programmatic verifiers.}
A programmatic verifier is any external procedure that can compute a target label, scalar, or accept/reject judgment for a model-produced claim. In this paper the verifier is not a human preference model; it is a checkable computation such as a theorem prover, a game engine, a factual-evidence label, or code execution. This matters because the verifier target can be attached to an inline natural-language claim and used to train or audit the hidden states of the rationale span.

\textbf{Lean and LeanCheck.}
Lean is an interactive theorem prover and functional programming language used to state mathematical theorems and mechanically check proofs \citep{demoura2021lean4}. A Lean proof either elaborates and is accepted by the kernel, or fails with an error. LeanCheck is our lightweight formal-verifier dataset built around this binary signal: each example contains a theorem, a candidate proof, a short informal rationale, and a verifier-derived label indicating whether Lean accepts the proof. The purpose is not to solve theorem proving end-to-end, but to test whether the selected text span encodes the formal accept/reject signal under controlled counterfactual swaps.

\textbf{KataGo and Go positions.}
KataGo is a strong open-source Go engine that combines neural-network evaluation with search in the AlphaGo/AlphaZero tradition \citep{silver2016mastering,silver2017mastering,wu2019accelerating}. For a board position it can estimate quantities such as win rate, score lead, ownership, and principal variations. We use KataGo as a dense programmatic verifier: instead of a single binary proof label, it supplies structured numeric or categorical targets for complex Go states. This makes the setting useful for testing whether natural-language commentary can encode verifier-derived information in a high-dimensional domain.

\section{Method}

The method has three components: a \emph{sequence format} that places a natural-language rationale and an explicit inline claim in the model's output, a \emph{consistency objective} that trains rationale-span hidden states to predict the programmatic verifier's label, and an optional \emph{information-flow constraint} (the BEC attention mask) that structurally prevents the claim head from bypassing the rationale. We then define a \emph{diagnostic ladder} that escalates from decodability through counterfactual following to causal influence. The following subsections define each component; implementation details (models, training) appear at the end of the consistency-objective subsection.

\subsection{Verifier-coupled sequence format}

For an input problem or state \(x\), a model generates a rationale \(r\) and an inline claim \(c\) in the format \texttt{[CONTEXT] x [RAT] r [CLAIM] c [EOS]}. A programmatic verifier \(V\) supplies a target \(y = V(x)\), such as a proof-checker accept/reject label, a win-rate bucket, a complexity class, or a factual evidence label. The inline claim is the text location where the model commits to this verifier-grounded statement.

\begin{figure*}[t]
    \centering
    \begin{tikzpicture}[
        scale=0.80, transform shape,
        font=\small,
        box/.style={rounded corners=4pt, draw=vcGray, very thick, align=center, inner sep=6pt, minimum height=0.78cm},
        mainbox/.style={box, draw=vcTeal, fill=white, very thick, minimum width=6.0cm, minimum height=1.35cm},
        auxbox/.style={box, draw=vcTeal, fill=vcSoft, minimum width=3.5cm},
        claimbox/.style={box, draw=vcGold, fill=vcWarm, minimum width=3.4cm},
        arrow/.style={-{Latex[length=2.2mm]}, very thick, draw=vcTeal},
        garrow/.style={-{Latex[length=2.2mm]}, very thick, draw=vcGold}
    ]
        \node[box, minimum width=2.55cm] (x) {Problem / state\\$x$};
        \node[mainbox, right=1.0cm of x] (seq) {
            \texttt{[CONTEXT]} $x$\\
            \texttt{[RAT]} natural-language rationale $r$\\
            \texttt{[CLAIM]} inline verifier claim $c$
        };
        \node[box, minimum width=2.9cm, right=1.0cm of seq] (verifier) {Programmatic\\verifier $V(x)$};
        \node[auxbox, below=0.82cm of seq] (pool) {Pool hidden states\\over selected span};
        \node[auxbox, below=0.58cm of pool] (head) {Consistency head\\predicts verifier label $y$};
        \node[claimbox, right=0.75cm of pool] (claimloss) {Claim loss\\$c$ vs. verifier $y$};

        \draw[arrow] (x) -- (seq);
        \draw[arrow] (seq) -- (verifier);
        \draw[arrow] (seq.south) -- (pool.north);
        \draw[arrow] (pool.south) -- (head.north);
        \draw[garrow] ($(verifier.south)+(0,-0.08)$) to[out=-82,in=18,looseness=0.85] ($(claimloss.east)+(0.08,0.12)$);
        \draw[garrow] (verifier.north west) to[out=165,in=15,looseness=1.0] (seq.north east);
        \draw[garrow] (head.east) to[out=10,in=230] (claimloss.south west);

        \node[below=0.28cm of head, align=center, text=vcText] (loss) {
            \(\displaystyle L = L_{\mathrm{LM}} + \lambda_{\mathrm{claim}}L_{\mathrm{claim}} + \lambda_{\mathrm{cons}}L_{\mathrm{cons}}\)
        };
        \node[below=0.12cm of loss, align=center, text=black!60, font=\footnotesize] {
            Consistency-head accuracy measures decodability; counterfactual swaps and activation patching test stronger forms of coupling.
        };
    \end{tikzpicture}
    \caption{\textbf{Verifier-coupled reasoning.} A model generates a rationale and inline claim. A programmatic verifier supplies the target label. The consistency head predicts that label from pooled hidden states over the selected text span, making rationale-to-claim coupling directly measurable.}
    \label{fig:method}
\end{figure*}

\subsection{Consistency objective}

Let \(h_i\) be the model hidden states and \(S_{\mathrm{rat}}\) the token span between \texttt{[RAT]} and \texttt{[CLAIM]}. A pooling operator produces \(z_{\mathrm{rat}} = \mathrm{Pool}(\{h_i : i \in S_{\mathrm{rat}}\})\), and a consistency head \(g\) predicts \(\hat{y}_{\mathrm{cons}} = g(z_{\mathrm{rat}})\). Training uses \(L = L_{\mathrm{LM}} + \lambda_{\mathrm{claim}} L_{\mathrm{claim}} + \lambda_{\mathrm{cons}} L_{\mathrm{cons}}\), where \(L_{\mathrm{LM}}\) is next-token prediction, \(L_{\mathrm{claim}}\) supervises the inline claim when applicable, and \(L_{\mathrm{cons}}\) predicts the verifier target from the selected span. Different variants change the pooling span: rationale-only, proof-only, full-sequence, claim-only, wrong-span, or random-label controls.

In practice, $g$ is a single linear layer ($\mathrm{Linear}(d_{\mathrm{model}}, |\mathcal{Y}|)$ where $|\mathcal{Y}|$ is the number of verifier label classes), and $\mathrm{Pool}$ is mean pooling over the span tokens. The base model is either a small decoder-only transformer trained from scratch (synthetic, KataGo, and code experiments; 2--4 layers, $d_{\mathrm{model}} \in \{64, 256\}$) or a pretrained GPT-2 fine-tuned on the task data (LeanCheck and FEVER experiments). Training uses AdamW on the combined loss; no reinforcement learning is applied. Full hyperparameters are in \Cref{app:repro}.

\subsection{Information-flow constraints}

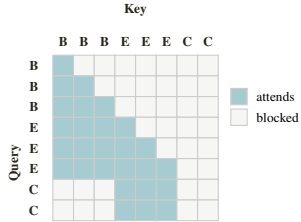
\begin{figure}[H]
\setlength{\fboxsep}{0pt}
\centering
\begin{tikzpicture}[
    scale=0.72, transform shape,
    font=\scriptsize,
    cell/.style={minimum size=0.38cm, inner sep=0pt, outer sep=0pt, draw=black!20, anchor=center},
    yes/.style={cell, fill=vcTeal!40},
    no/.style={cell, fill=vcGray!20},
    lbl/.style={font=\scriptsize\bfseries, text=vcText}
]
    \def\s{0.38}

    % Column labels (keys)
    \node[lbl] at ({0*\s+0.5*\s}, {8*\s+0.25}) {B};
    \node[lbl] at ({1*\s+0.5*\s}, {8*\s+0.25}) {B};
    \node[lbl] at ({2*\s+0.5*\s}, {8*\s+0.25}) {B};
    \node[lbl] at ({3*\s+0.5*\s}, {8*\s+0.25}) {E};
    \node[lbl] at ({4*\s+0.5*\s}, {8*\s+0.25}) {E};
    \node[lbl] at ({5*\s+0.5*\s}, {8*\s+0.25}) {E};
    \node[lbl] at ({6*\s+0.5*\s}, {8*\s+0.25}) {C};
    \node[lbl] at ({7*\s+0.5*\s}, {8*\s+0.25}) {C};
    \node[lbl, anchor=south] at ({4*\s}, {8*\s+0.6}) {Key};

    % Row labels (queries)
    \node[lbl] at ({-0.35}, {7*\s+0.5*\s}) {B};
    \node[lbl] at ({-0.35}, {6*\s+0.5*\s}) {B};
    \node[lbl] at ({-0.35}, {5*\s+0.5*\s}) {B};
    \node[lbl] at ({-0.35}, {4*\s+0.5*\s}) {E};
    \node[lbl] at ({-0.35}, {3*\s+0.5*\s}) {E};
    \node[lbl] at ({-0.35}, {2*\s+0.5*\s}) {E};
    \node[lbl] at ({-0.35}, {1*\s+0.5*\s}) {C};
    \node[lbl] at ({-0.35}, {0*\s+0.5*\s}) {C};
    \node[lbl, anchor=east, rotate=90] at ({-0.7}, {4*\s}) {Query};

    % Board rows (0-2): attend to preceding Board tokens (causal)
    \foreach \i in {0,...,2} {
        \foreach \j in {0,...,7} {
            \pgfmathparse{int((\j<=\i) ? 1 : 0)}
            \ifnum\pgfmathresult=1
                \node[yes] at ({\j*\s+0.5*\s}, {(7-\i)*\s+0.5*\s}) {};
            \else
                \node[no] at ({\j*\s+0.5*\s}, {(7-\i)*\s+0.5*\s}) {};
            \fi
        }
    }
    % Explanation rows (3-5): attend to all Board + causal Expl
    \foreach \i in {3,...,5} {
        \foreach \j in {0,...,7} {
            \pgfmathparse{int((\j<=2 || (\j>=3 && \j<=\i)) ? 1 : 0)}
            \ifnum\pgfmathresult=1
                \node[yes] at ({\j*\s+0.5*\s}, {(7-\i)*\s+0.5*\s}) {};
            \else
                \node[no] at ({\j*\s+0.5*\s}, {(7-\i)*\s+0.5*\s}) {};
            \fi
        }
    }
    % Claim rows (6-7): attend ONLY to Explanation tokens
    \foreach \i in {6,...,7} {
        \foreach \j in {0,...,7} {
            \pgfmathparse{int((\j>=3 && \j<=5) ? 1 : 0)}
            \ifnum\pgfmathresult=1
                \node[yes] at ({\j*\s+0.5*\s}, {(7-\i)*\s+0.5*\s}) {};
            \else
                \node[no] at ({\j*\s+0.5*\s}, {(7-\i)*\s+0.5*\s}) {};
            \fi
        }
    }

    % Legend
    \node[yes, minimum size=0.3cm] at ({9*\s}, {6*\s}) {};
    \node[anchor=west, font=\scriptsize] at ({9*\s+0.2}, {6*\s}) {attends};
    \node[no, minimum size=0.3cm] at ({9*\s}, {5*\s}) {};
    \node[anchor=west, font=\scriptsize] at ({9*\s+0.2}, {5*\s}) {blocked};
\end{tikzpicture}
\caption{\textbf{BEC attention mask.} Board (B) tokens use standard causal attention. Explanation (E) tokens attend to all Board tokens and preceding Explanation tokens. Claim (C) tokens attend \emph{only} to Explanation tokens, structurally preventing the claim head from bypassing the explanation.}
\label{fig:bec}
\end{figure}

An optional architectural complement to the consistency objective is a structured attention mask that constrains which tokens each output can attend to. In the Board-Explanation-Claim (BEC) variant, the input sequence is divided into three blocks: board or context tokens (B), explanation tokens (E), and claim tokens (C). Board tokens use standard causal self-attention. Explanation tokens attend to all board tokens and to preceding explanation tokens. Claim tokens attend \emph{only} to a windowed subset of explanation tokens---not to board tokens and not to other claims. This makes it structurally impossible for the claim head to bypass the explanation and read verifier-relevant information directly from the input. The BEC mask is implemented as an additive attention bias and is used in the Go GPT-OSS experiments reported in \Cref{app:tables}. \Cref{fig:bec} illustrates the attention pattern.

\subsection{Diagnostic ladder}

The central distinction is between \emph{decodability} and \emph{faithfulness}. The consistency head measures whether verifier information can be decoded from rationale states. Stronger diagnostics ask whether the representation follows counterfactual swaps and whether intervening on it changes claim predictions.

\begin{table}[t]
\vspace{0.2em}
\caption{Diagnostics used to interpret verifier-coupled reasoning.}
\label{tab:diagnostics}
\centering
\scriptsize
\setlength{\tabcolsep}{3pt}
\renewcommand{\arraystretch}{0.88}
\begin{tabularx}{\columnwidth}{@{}p{0.20\columnwidth}XX@{}}
\toprule
Diagnostic & Supports & Caveat \\
\midrule
Claim acc. & Correct claim & Rationale may be post-hoc \\
Head acc. & Label decodable & Not causal by itself \\
Swap & Span-specific following & May exploit templates \\
Controls & Not just capacity & Not sufficient alone \\
Patching & Causal influence & Needs careful baselines \\
\bottomrule
\end{tabularx}
\end{table}

\section{Experiments}

Each setting tests a different facet of the decodability--faithfulness gap. Synthetic experiments isolate the mechanism and provide causal evidence via activation patching. LeanCheck tests formal-verifier coupling with counterfactual span separation. KataGo extends coupling to dense, non-binary verifier labels. Code coupling reveals the central failure mode: near-perfect decodability with zero generation quality. FEVER adds a real-language setting with an accuracy--sensitivity trade-off. Together, these settings trace the diagnostic ladder: each probes a progressively stronger condition, and each reveals limitations the previous level misses. Detailed per-variant tables are in \Cref{app:tables}.

\subsection{Synthetic: controlled coupling and interventions}

The synthetic setting is the cleanest place to isolate the mechanism. Each example contains a latent state, a natural-language rationale, and a deterministic claim label. Early runs with eight latent states showed the basic effect: without consistency loss, the classifier decoding claim labels from rationale states was near zero (3.9\%) even when generation accuracy reached 75.0\%; consistency-trained variants reached 100.0\% classifier accuracy, with generation accuracy varying by pooling span. This already suggested that the LM objective can learn to emit claims without making rationale states predictive of them.

The stronger synthetic result comes from scaled convergence and intervention runs. At convergence, all main variants reached 100\% generation accuracy, but the baseline stopped reading rationales: it could generate the right claim while its rationale representation did not carry the relevant label signal. By contrast, consistency-trained models preserved rationale-to-claim coupling. A claim-only pooling negative control reached only 43\% classifier accuracy, showing that the result depends on the selected span rather than merely on model capacity. In a harder overlapping-vocabulary run, rationale templates shared roughly 50\% of token positions across latent states. Even there, \texttt{rationale\_only} and \texttt{earlier\_token\_only} reached 100\% generation accuracy, 100\% rationale-pooled classifier accuracy, and 100\% counterfactual classifier following; \texttt{full\_sequence} kept 100\% classifier coupling but had lower generation accuracy at 81.25\%. The no-consistency baseline still had 100\% generation accuracy but only 4.69\% rationale-pooled classifier accuracy and 6.25\% counterfactual classifier following.

Most importantly, hidden-state intervention tests provide causal evidence---testing whether rationale representations merely correlate with the claim (decodability) or actually determine it (causal influence), the central distinction in the diagnostic ladder. Patching rationale hidden states at layer 0 changed the downstream claim in 73--89\% of cases for consistency-trained variants, compared with 31\% for the baseline. At layer~1 (the final block), all variants show 0\% following of the patched states, confirming that the intervention must occur early enough for downstream blocks to propagate the signal. The full per-layer, per-variant breakdown is in \Cref{app:intervention}. This directly addresses the diagnostic ladder: consistency loss is not only making labels decodable; in this controlled setting, rationale activations causally affect claim behavior.

\subsection{LeanCheck: formal-verifier span separation}

LeanCheck instantiates verifier-coupled reasoning with a formal proof checker. Each example contains a Lean theorem, a candidate proof, a natural-language rationale, and an inline claim indicating whether Lean accepts or rejects the proof. The model is not asked to synthesize proofs; the task isolates whether informal rationale representations encode formal verifier outcomes. The run uses GPT-2 with 1,000 Lean-checked training examples, 200 evaluation examples, 200 counterfactual swaps, and 200 minimal-pair rows. The data covers natural-number equalities, propositional logic, and simple list lemmas, with negative examples generated by wrong lemmas, renamed variables, deleted proof lines, replacement tactics, missing premises, wrong pairings, and adversarial near misses.

Concretely, the data covers six Lean~4 theorem templates in three families: natural-number arithmetic (\texttt{Nat.add\_comm}, \texttt{Nat.add\_zero}), propositional logic (\texttt{exact hp}, \texttt{And.intro hp hq}), list algebra (\texttt{List.append\_nil}), and reflexive equality. Negative examples apply one of seven mutations to the correct proof. \emph{Wrong-lemma} substitutes a type-incorrect lemma---for example, \texttt{Nat.mul\_comm a b} for an addition-commutativity goal, producing a type mismatch because $a \cdot b = b \cdot a$ does not close $a + b = b + a$. \emph{Rename-variable} introduces an unbound name (\texttt{And.intro hp hr} when only \texttt{hp} and \texttt{hq} are in scope). \emph{Delete-final-line} empties the proof body, leaving unsolved goals. \emph{Replace-tactic} substitutes a plausible but insufficient tactic such as \texttt{rfl} for a non-trivially-equal goal. The remaining strategies---wrong-pairing, missing-premise, and adversarial-near-miss---cover cross-family confusion, removed hypotheses, and syntactically plausible proofs of subtly different propositions. All 1,000 training labels come from the Lean~4 kernel, not human annotation. Rationales are sampled from per-domain and per-mutation paraphrase pools (five negative paraphrases per mutation strategy, four positive paraphrases per domain) to discourage template memorization.

LeanCheck gives the cleanest span-specific result. Without consistency loss, classifier accuracy is near chance: \texttt{lm\_only} reaches 49.0\% and \texttt{no\_consistency\_loss} reaches 51.0\%. Wrong-span pooling is also near chance at 52.0\%, and random-label consistency is 41.0\%. In contrast, \texttt{rationale\_only}, \texttt{proof\_only}, and \texttt{full\_sequence} all reach 100.0\% classifier accuracy and 100.0\% minimal-pair flip accuracy.

The key result is the counterfactual split. Counterfactual examples are constructed by cross-wiring proof and rationale evidence. In one direction, a correct proof (e.g.\ \texttt{exact And.intro hp hq}, which Lean accepts) is paired with a negative rationale (``A proof of p alone does not prove the conjunction p and q'') and the label FAILS. In the mirror direction, a broken proof (\texttt{exact hp}, which Lean rejects because \texttt{hp} has type \texttt{p} rather than \texttt{p $\wedge$ q}) is paired with a positive rationale (``The candidate proof and theorem target align at the type-checker level'') and the label VERIFIES. The 200 counterfactual rows split evenly between these two swap directions, so both the rationale-pooled and proof-pooled heads face unambiguous label conflicts.

The \texttt{rationale\_only} head follows the swapped rationale label 100.0\% of the time and almost never follows the proof label (1.0\%). The \texttt{proof\_only} head shows the mirror image: it follows the proof label 100.0\% of the time and the rationale label only 1.0\%. This perfect separation is exactly what the mechanism predicts (\Cref{fig:leancheck-ex}). It shows that the selected text span determines which verifier signal the head follows.

Minimal pairs provide a complementary diagnostic. Each pair shares the same theorem with exactly one proof-line mutation: for example, \texttt{Nat.add\_comm a b} (accepted) versus \texttt{Nat.mul\_comm a b} (rejected, since multiplication commutativity cannot close an addition goal). All consistency-trained variants reach 100\% minimal-pair flip accuracy, confirming that even single-mutation differences are reliably reflected in the pooled representation. Controls (\texttt{lm\_only}, \texttt{no\_consistency\_loss}, \texttt{wrong\_span}) reach 0\%.

LeanCheck activation patching provides additional evidence. For \texttt{rationale\_only}, rationale-span patches produce a large positive head-logit effect (11.58) and a positive rationale-minus-random effect (4.52). Baselines do not show the same head-level pattern. LM-logit patching is noisier, but the head-patching results support the claim that trained rationale states influence the verifier classifier in the intended direction.

\begin{figure}[H]
\setlength{\fboxsep}{4pt}
\colorbox{vcSoft}{\begin{minipage}{0.94\columnwidth}\footnotesize
\textbf{Example: LeanCheck counterfactual pair.}
Both examples use the same theorem: \texttt{example (p q : Prop) (hp : p) (hq : q) : p $\wedge$ q := by}. The rationales are cross-swapped so that proof and rationale labels conflict.

\medskip
\textbf{(a)} Proof: \texttt{exact And.intro hp hq} --- Lean accepts.\\
Rationale: ``A proof of p alone does not prove the conjunction p and q.''\\
Rationale-pooled head predicts \textbf{FAILS} (follows rationale, not proof).

\medskip
\textbf{(b)} Proof: \texttt{exact hp} --- Lean rejects (\texttt{hp} has type \texttt{p} but expected \texttt{p $\wedge$ q}).\\
Rationale: ``The candidate proof and theorem target align at the type-checker level.''\\
Rationale-pooled head predicts \textbf{VERIFIES} (follows rationale, not proof).

\medskip
{\footnotesize The rationale-pooled head follows the rationale in both directions; the proof-pooled head does the mirror, following the proof 100\% of the time.}
\end{minipage}}
\caption{LeanCheck counterfactual pair from the evaluation data. The consistency head follows whichever span it pools over, producing perfect directional separation under label conflict.}
\label{fig:leancheck-ex}
\end{figure}

\subsection{KataGo: dense domain verifier coupling}

KataGo provides dense programmatic labels for Go positions, building on the neural-network and tree-search paradigm behind superhuman Go systems \citep{silver2016mastering,silver2017mastering,wu2019accelerating}. The dataset contains 3,748 positions from 954 reviewed games (3,373 train, 375 eval), predominantly 19$\times$19 boards. Each position is linearized into a symbolic token sequence: metadata tokens for board size, side to move, komi, and ruleset, followed by stone-coordinate tokens such as \texttt{B\_Q4} or \texttt{W\_D17}, sorted by color and then alphabetically. A natural-language commentary span follows the position tokens, and the sequence ends with an inline claim token encoding the KataGo-derived win-rate bin (V0--V9, each spanning roughly 10 percentage points). The win-rate distribution is U-shaped: bins V0 and V9 (decisive positions) account for 56\% of the data, reflecting the natural distribution of positions sampled from real games (\Cref{fig:katago-ex}).

The model is a 4-layer decoder-only transformer ($d_{\mathrm{model}} = 256$, 8 heads, $d_{\mathrm{ff}} = 1024$) trained from scratch with three output heads sharing the same hidden states: a language-modeling head over the full vocabulary, a scalar head that reads the hidden state at the claim-token position and predicts continuous win probability via sigmoid, and a consistency head that mean-pools hidden states over the commentary span and classifies the win-rate bin. The five ablations vary which heads receive gradient: \texttt{lm\_only} trains only the LM head; \texttt{no\_consistency\_loss} adds the scalar head; \texttt{rationale\_only} adds only the bin-classification head; \texttt{full\_consistency} trains all three; and \texttt{random\_consistency} trains the bin head on shuffled labels as a control.

This setting matters because it moves beyond binary accept/reject labels. Go commentary requires mapping a complex board state into qualitative natural-language explanation while grounding the claim in an external engine. The result is strong: \texttt{lm\_only} reaches 1.6\% claim-bin accuracy and \texttt{no\_consistency\_loss} reaches 25.1\%. With consistency training, \texttt{rationale\_only} reaches 78.7\% and \texttt{full\_consistency} reaches 81.1\%, while \texttt{random\_consistency} remains low at 16.0\%. Notably, \texttt{full\_consistency} also achieves near-perfect scalar regression (Pearson $r = 0.994$, MAE $= 0.028$), showing that the bin-classification objective does not interfere with fine-grained win-rate prediction. On counterfactual swaps, the consistency-trained variants follow the swapped commentary label 52--55\% of the time versus only 14.5\% following the original position's label, while baselines show no such asymmetry. This confirms that natural-language commentary states can be trained to encode dense verifier-derived labels, not only toy latent states or formal accept/reject signals.

\begin{figure}[H]
\setlength{\fboxsep}{4pt}
\colorbox{vcSoft}{\begin{minipage}{0.94\columnwidth}\footnotesize
\textbf{Example: KataGo dense verifier coupling.}
Position: 19$\times$19, move~5, White to play, komi~6.5. Black: Q15, Q10, Q5. White: D17, D3.

\medskip
\textbf{Commentary:} ``[\,\ldots\,] I would prefer to be white here. To me the 4th line gives influence but the tight placement compared to san-ren-sei is territory focussed. The structure is very easy to reduce with both the 3-3 and 3-4 points open so there is no urgency for white to reduce.''

\medskip
\textbf{KataGo verifier targets:} win rate 29.2\% Black (bin~3/9), score lead $-1.86$, phase: opening, urgency: medium, best move: R3.

\medskip
{\footnotesize The consistency head predicts the win-rate bin from commentary-span hidden states. With consistency training, claim-bin accuracy reaches 81.1\% vs.\ 25.1\% without.}
\end{minipage}}
\caption{KataGo example from the training data. Natural-language Go commentary is paired with dense engine-derived targets; the model must encode the win-rate bucket from the commentary span.}
\label{fig:katago-ex}
\end{figure}

\subsection{Code: representation coupling succeeds, faithful generation does not}

The code experiment asks whether the mechanism generalizes to algorithmic claims such as time complexity, space complexity, algorithm class, loop structure, key operation, access pattern, auxiliary structures, input mutation, correctness status, and edge-case handling.

\textbf{Coupling succeeds.} In the standard three-claim ablation, the full consistency-loss variant reaches 100\% coupling strength, 100\% counterfactual swap influence, and 100\% structured claim-tag accuracy, while \texttt{no\_consistency\_loss} reaches only 17.5\% coupling and 5\% swap influence. In the richer 12-claim ontology, \texttt{consistency\_loss} reaches .986 mean coupling and about .978 swap influence, compared with .286 and $-.033$ for \texttt{no\_consistency\_loss}. The per-claim breakdown reveals an asymmetry: binary and low-cardinality claims are partially learnable from code structure alone (\texttt{mutates\_input} reaches 77.4\% even \emph{without} consistency loss), whereas fine-grained claims are where the objective matters most: \texttt{algorithm\_class} reaches 99.0\% with consistency loss versus 13.4\% without, and \texttt{key\_operation} 99.8\% versus 9.6\% (\Cref{tab:perclaim}). Representation coupling is therefore won decisively.

\begin{figure}[t]
\setlength{\fboxsep}{4pt}
\colorbox{vcWarm}{\begin{minipage}{0.94\columnwidth}\footnotesize
\textbf{Example: coupling succeeds, the explanation is fluent but unfaithful.}
Input: \texttt{selection\_sort}, a standard nested-loop sort that finds and swaps the minimum element into position.

\medskip
\textbf{Oracle claims:} time $O(n^2)$, space $O(1)$, correct.\\
\textbf{Model structured claims:} time $O(n^2)$, space $O(1)$, correct, all match the oracle.

\medskip
\textbf{Model prose (verbatim):} ``Computes the sum of a list with a single pass.\ Time complexity O(n), space O(1).''

\medskip
{\footnotesize The prose is fluent and grammatical but describes a different algorithm (``sum of a list'') with the wrong time complexity ($O(n)$ rather than $O(n^2)$), while the structured claim tags correctly emit $O(n^2)$. This is representative: in the controlled re-run with a standard subword tokenizer (\Cref{sec:ceilings}), generated explanations state the correct time complexity in only ${\sim}18\%$ of cases, below the 33\% three-class chance rate, and never reproduce the code's reference explanation, identically for pretrained and from-scratch GPT-2. The failure mode is infidelity, not incoherence.}
\end{minipage}}
\caption{Code failure mode. The model's hidden states encode the verifier-derived claims and it emits the correct structured claim tags, but its generated prose, though fluent, describes an unrelated algorithm. The text shown is verbatim model output.}
\label{fig:code-ex}
\end{figure}

\textbf{Faithful generation does not.} Winning coupling does not yield a faithful explanation. We assess generation in a controlled run that decodes with a standard subword tokenizer (GPT-2, both pretrained and trained from scratch; \Cref{sec:ceilings}). The generated prose is fluent and grammatical (no degeneration, no repetition), but it is \emph{unfaithful}: it describes a different algorithm than the input code. Given \texttt{selection\_sort}, the model emits the correct structured claims ($O(n^2)$, $O(1)$, correct) while its prose reads ``Computes the sum of a list with a single pass'' and asserts the wrong complexity (\Cref{fig:code-ex}). Across the reviewed examples the generated explanation states the correct time complexity in only about 18\% of cases, below the 33\% three-class chance rate, and never reproduces the code's reference explanation; pretrained and from-scratch models behave identically. The code result is thus neither a coupling failure nor an incoherence failure: it shows that near-perfect coupling metrics can be won while the generated explanation remains unfaithful to the code it accompanies.

\textbf{Training dynamics.} The two outcomes separate cleanly over training. Coupling strength reaches 98.6\% by epoch~5 for the consistency-loss variant and stays there; the no-consistency baseline never exceeds 29\% across 20~epochs (\Cref{tab:code_dynamics}). Faithful generation, by contrast, never emerges at any epoch, because nothing in the training objective rewards the \emph{generated text} for matching the verifier: the consistency loss shapes the rationale representation, while the language-modeling loss is trained on explanations that are themselves not tied to the code (\Cref{sec:ceilings}). Coupling is learned quickly and stably; faithful explanation is not learned at all.

\subsection{FEVER: evidence sensitivity in real language}

FEVER tests verifier-coupled reasoning on real-language evidence verification \citep{thorne2018fever}. The model sees evidence text, a claim sentence, and an inline label token (SUPPORTS, REFUTES, or NOT ENOUGH INFO). The key diagnostic holds the claim text fixed while swapping evidence, isolating whether the model reads evidence or defaults to claim-text shortcuts. We report results from a pretrained GPT-2 run on 50k train / 5k eval examples and a from-scratch 4-layer transformer on the same split.

Evidence-only pooling has lower raw classifier accuracy than claim-visible pooling but better isolates evidence sensitivity. In the pretrained run, \texttt{evidence\_only\_pooling} reaches about 44\% classifier accuracy and 81\% generation accuracy; on matched counterfactuals, it follows swapped evidence about 46\% of the time and the original label about 23\%. Claim-visible pooling (\texttt{full\_sequence}, \texttt{claim\_only}) reaches about 84\% raw classifier accuracy but shows more inertia toward the original claim, consistent with leveraging the claim text rather than parsing the evidence. Random-label evidence pooling falls near chance at about 23\%.

The from-scratch run makes the gap more dramatic (\Cref{tab:fever_scratch_main}): \texttt{full\_sequence\_pooling} reaches 99.0\% classifier accuracy and 97.2\% counterfactual swap following---near-perfect coupling when the claim text is visible---while evidence-only pooling stays at 44\% classifier accuracy and 39\% swap following. This contrast shows that a model trained from scratch can learn the claim-label correlation almost perfectly while still struggling to extract the same signal from evidence text alone. The evidence-only difficulty is not a pretrained shortcut artifact; it reflects a genuine asymmetry in how much verifier-relevant information is accessible from different text spans.

\begin{table}[t]
\caption{FEVER from-scratch run. Full-sequence pooling achieves near-perfect coupling while evidence-only pooling stays near 44\%, revealing an accuracy-sensitivity trade-off that final-answer accuracy would hide.}
\label{tab:fever_scratch_main}
\centering
\small
\setlength{\tabcolsep}{4pt}
\renewcommand{\arraystretch}{0.9}
\begin{tabular}{lrrrr}
\toprule
Variant & Cls & Gen & Swap & Orig \\
\midrule
\texttt{full\_sequence} & .990 & .972 & .972 & .014 \\
\texttt{claim\_only} & .984 & .966 & .960 & .022 \\
\texttt{evidence\_only} & .440 & .812 & .390 & .320 \\
\texttt{no\_consistency} & .498 & .806 & .340 & .360 \\
\bottomrule
\end{tabular}
\vspace{-0.3em}
\end{table}

FEVER thus contributes a distinct lesson: in natural-language settings, the span that provides the strongest raw accuracy (claim text) may not be the span that tests whether the model actually processes the evidence. The evidence-only constraint exposes an accuracy-sensitivity trade-off invisible to generation-level evaluation.

\begin{table}[t]
\caption{Main experimental arc. Metrics are setting-specific primary diagnostics and should be compared within settings; acc. denotes accuracy.}
\label{tab:summary}
\centering
\small
\setlength{\tabcolsep}{4pt}
\renewcommand{\arraystretch}{0.9}
\begin{tabular}{llll}
\toprule
Setting & Diagnostic & Trained & Control \\
\midrule
Synthetic & layer-0 patch & 73--89\% & 31\% \\
LeanCheck & span follows & 100\% & 50--52\% \\
KataGo & claim-bin acc. & 81.1\% & 25.1\% \\
Code & mean coupling & 98.6\% & 28.6\% \\
FEVER & evid.\ swap fol. & 46\% & 34\% \\
\bottomrule
\end{tabular}
\vspace{0.3em}
\end{table}

\begin{figure*}[t]
    \centering
    \begin{tikzpicture}[x=1cm,y=0.017cm,font=\scriptsize, scale=0.62, transform shape]
        \def\panelw{2.55}
        \def\barw{0.38}
        \def\gap{0.68}

        % shared y-axis and grid
        \foreach \y in {0,20,40,60,80,100} {
            \draw[black!10] (0,\y) -- (15.35,\y);
            \node[anchor=east, black!55] at (-0.08,\y) {\y};
        }
        \node[rotate=90, anchor=south, black!75] at (-0.65,50) {Primary diagnostic (\%)};

        % Panel 1: Synthetic
        \begin{scope}[shift={(0,0)}]
            \node[align=center,font=\small\bfseries] at (1.15,135) {Synthetic\\layer-0 patch};
            \fill[vcGray] (0.15,0) rectangle +(0.38,31);
            \fill[vcRust] (0.83,0) rectangle +(0.38,73);
            \fill[vcTeal] (1.51,0) rectangle +(0.38,89);
            \node[font=\scriptsize\bfseries] at (0.34,35) {31\%};
            \node[font=\scriptsize\bfseries] at (1.02,77) {73\%};
            \node[font=\scriptsize\bfseries] at (1.70,93) {89\%};
            \node[rotate=30,anchor=east] at (0.52,-8) {Baseline};
            \node[rotate=30,anchor=east] at (1.20,-8) {Cons. low};
            \node[rotate=30,anchor=east] at (1.88,-8) {Cons. high};
        \end{scope}

        % Panel 2: LeanCheck
        \begin{scope}[shift={(3.25,0)}]
            \node[align=center,font=\small\bfseries] at (1.15,135) {LeanCheck\\directional cfact};
            \fill[vcGray] (0.15,0) rectangle +(0.38,50.5);
            \fill[vcRust] (0.83,0) rectangle +(0.38,100);
            \fill[vcTeal] (1.51,0) rectangle +(0.38,100);
            \node[font=\scriptsize\bfseries] at (0.34,54.5) {50\%};
            \node[font=\scriptsize\bfseries, text=black] at (1.02,104) {100\%};
            \node[font=\scriptsize\bfseries, text=black] at (1.70,104) {100\%};
            \node[rotate=30,anchor=east] at (0.52,-8) {No cons.};
            \node[rotate=30,anchor=east] at (1.20,-8) {Rat. head};
            \node[rotate=30,anchor=east] at (1.88,-8) {Proof head};
        \end{scope}

        % Panel 3: KataGo
        \begin{scope}[shift={(6.50,0)}]
            \node[align=center,font=\small\bfseries] at (1.15,135) {KataGo\\claim-bin acc.};
            \fill[vcGray] (0.15,0) rectangle +(0.38,25.1);
            \fill[vcRust] (0.83,0) rectangle +(0.38,78.7);
            \fill[vcTeal] (1.51,0) rectangle +(0.38,81.1);
            \node[font=\scriptsize\bfseries] at (0.34,29.1) {25\%};
            \node[font=\scriptsize\bfseries] at (1.02,82.7) {79\%};
            \node[font=\scriptsize\bfseries] at (1.70,85.1) {81\%};
            \node[rotate=30,anchor=east] at (0.52,-8) {No cons.};
            \node[rotate=30,anchor=east] at (1.20,-8) {Rationale};
            \node[rotate=30,anchor=east] at (1.88,-8) {Full cons.};
        \end{scope}

        % Panel 4: Code
        \begin{scope}[shift={(9.75,0)}]
            \node[align=center,font=\small\bfseries] at (1.15,135) {Code\\mean coupling};
            \fill[vcGray] (0.15,0) rectangle +(0.38,28.6);
            \fill[vcRust] (0.83,0) rectangle +(0.38,56.2);
            \fill[vcTeal] (1.51,0) rectangle +(0.38,98.6);
            \node[font=\scriptsize\bfseries] at (0.34,32.6) {29\%};
            \node[font=\scriptsize\bfseries] at (1.02,60.2) {56\%};
            \node[font=\scriptsize\bfseries, text=black] at (1.70,102.6) {99\%};
            \node[rotate=30,anchor=east] at (0.52,-8) {No cons.};
            \node[rotate=30,anchor=east] at (1.20,-8) {Random};
            \node[rotate=30,anchor=east] at (1.88,-8) {Cons.};
        \end{scope}

        % Panel 5: FEVER
        \begin{scope}[shift={(13.00,0)}]
            \node[align=center,font=\small\bfseries] at (1.15,135) {FEVER\\evid.\ swap fol.};
            \fill[vcGray] (0.15,0) rectangle +(0.38,34);
            \fill[vcRust] (0.83,0) rectangle +(0.38,46);
            \fill[vcTeal] (1.51,0) rectangle +(0.38,97.2);
            \node[font=\scriptsize\bfseries] at (0.34,38) {34\%};
            \node[font=\scriptsize\bfseries] at (1.02,50) {46\%};
            \node[font=\scriptsize\bfseries, text=black] at (1.70,101.2) {97\%};
            \node[rotate=30,anchor=east] at (0.52,-8) {No cons.};
            \node[rotate=30,anchor=east] at (1.20,-8) {Evid.\ only};
            \node[rotate=30,anchor=east] at (1.88,-8) {Full seq.};
        \end{scope}

        \draw[black!25] (0,0) -- (15.35,0);
        \draw[black!25] (0,0) -- (0,105);
    \end{tikzpicture}
    \caption{{\footnotesize \textbf{Primary coupling diagnostics across all five settings.} Metrics are setting-specific: synthetic layer-0 intervention effect, LeanCheck directional counterfactual (cfact) following, KataGo claim-bin accuracy (acc.), code mean coupling, and FEVER evidence-swap following. In LeanCheck, the two 100\% bars are different directional tests: the rationale-pooled head follows the rationale label, while the proof-pooled head follows the proof label. FEVER shows the accuracy-sensitivity trade-off: evidence-only pooling (46\%) has lower raw accuracy than full-sequence (97.2\%) but better isolates evidence sensitivity. Values should be compared within each setting, not as one universal score.}}
    \label{fig:bars}
\end{figure*}
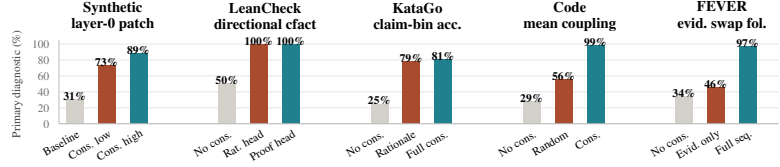

\section{Discussion}

\subsection{What the settings jointly show}

The cross-setting pattern is now clearer. Synthetic experiments show that consistency loss can prevent the LM objective from bypassing rationale states and can create causal intervention effects. LeanCheck shows that the mechanism works in a formal-verifier setting and that the selected pooling span determines whether the classifier follows the rationale or proof under counterfactual conflict. KataGo shows that the same idea scales to dense, non-binary verifier labels in a complex domain. Code shows the key failure mode: representation coupling can be nearly perfect while generated explanations are unusable. FEVER contributes a complementary lesson about real-language settings: the span that maximizes raw accuracy (claim text at 99\%) may not be the span that tests whether the model actually processes the evidence (evidence-only at 44\%).

This distinction matters for mathematical and scientific agents. If the goal is to audit reasoning, a high consistency-head score is useful: it tells us that verifier-relevant information is present in the selected representation. If the goal is to trust the generated explanation, decodability alone does not suffice---the explanation must also be fluent, semantically grounded, and causally connected to the emitted claim. This is what the diagnostic ladder is designed to test.

\subsection{The decodability--faithfulness gap is structural}
\label{sec:ceilings}

The experiments establish that supervised consistency training reliably produces \emph{decodability} (verifier labels become linearly recoverable from rationale-span hidden states, reaching near-perfect coupling in the code setting) but that decodability does not, on its own, yield \emph{faithful generation}. Two limits recur. First, span selection trades diagnostic power against predictive accuracy: the span that most cleanly isolates verifier sensitivity is not the span that maximizes raw accuracy. Second, and more fundamentally, the consistency objective shapes the rationale \emph{representation} without constraining the generated \emph{text}. One might worry that the code experiment's gap is incidental to the objective and instead reflects a capacity-starved from-scratch model, or a too-low loss weight. We report two controlled experiments showing it is neither.

\textbf{The gap is not capacity-driven.} We hold the code-coupling task, dataset, and consistency objective fixed and train two models that differ \emph{only} in initialization: a pretrained GPT-2 (124M) and a from-scratch GPT-2 of identical architecture and subword tokenizer, three seeds each (\Cref{tab:capacity}). If unfaithful generation were a capacity artifact, generative pretraining should narrow the gap. It does not: both conditions reach identical coupling ($1.00 \pm 0.00$) and statistically identical BLEU-1 ($0.191 \pm 0.017$ vs.\ $0.183 \pm 0.047$; 4-gram repetition $0.000$ in either); pretraining only accelerates convergence (epoch-1 LM loss $1.19$ vs.\ $2.59$). The gap is therefore a property of the \emph{objective}, not of model scale or generative competence. The comparison also yields a methodological correction: under a standard subword tokenizer the generated prose is fluent in \emph{both} conditions, so the operative failure is not incoherence but \emph{infidelity}, i.e.\ the model emits a well-formed explanation of an unrelated algorithm.

\textbf{Reweighting the loss cannot close the gap.} Could up-weighting $L_{\mathrm{cons}}$ relative to $L_{\mathrm{LM}}$ pull generation toward faithfulness? It cannot, for a structural reason: in $L = L_{\mathrm{LM}} + \lambda_{\mathrm{cons}} L_{\mathrm{cons}}$, $L_{\mathrm{cons}}$ is a loss on a \emph{representation}, a probe over pooled hidden states, and only $L_{\mathrm{LM}}$ has a gradient path to emitted tokens. Sweeping $\lambda_{\mathrm{cons}}$ from $1$ to $10^{4}$ (\Cref{tab:lambda}) leaves every generation metric unchanged: coupling stays at $1.00$ and prose quality stays flat. Coupling reaches $1.00$ within the first epoch, after which $L_{\mathrm{cons}}$ and its gradient collapse to zero, so $\lambda_{\mathrm{cons}}$ multiplies a vanished gradient regardless of magnitude, and the final LM loss is identical across the entire sweep. A loss on the representation has no lever on generation, and no choice of weight converts it into one.

\textbf{Scope of the faithfulness claim.} We use ``unfaithful'' here in the operational sense that the generated prose and the inline structured claim refer to different algorithms; we do not equate it with faithfulness in the full causal sense of \citet{jacovi2020faithful} or \citet{lanham2023measuring}, which would require showing the explanation reflects the causal computation behind the answer (e.g.\ generation-side activation patching on a fluent pretrained model). Our claim is the weaker but substantive one that decodable verifier information in rationale hidden states is not sufficient for the generated explanation to refer to the same object as its verified claim. One design choice interacts with this scope: the LM head is trained on $(\textsc{code}, \textsc{mismatched-explanation})$ pairs by construction, so the prose is unfaithful partly by training-data design. The structural finding survives this caveat, since a representation-side loss cannot, even at $\lambda_{\mathrm{cons}}=10^{4}$, pull generation back toward the verified claim; a fully causal account in the Lanham/Jacovi sense will require generation-side interventions we leave to future work.

\textbf{What closing the gap requires.} Together these results show that the gap is structural: consistency training is an effective \emph{diagnostic} (it measures whether verifier-relevant information is present in the rationale representation, a necessary precondition for faithfulness) and an effective representation-shaping prior, but it is not, and cannot by reweighting become, a generation objective. Closing the gap requires an objective whose gradient reaches the generated tokens and rewards them for verifier consistency. The natural instrument is reinforcement learning with programmatic verifier rewards. Inline claims create checkable units: the model generates a rationale and an inline claim, the verifier scores the emitted claim, and the policy is rewarded or penalized on what it actually produced rather than on an auxiliary probe. Such RL should be paired with the intervention-based evaluation developed here, so that optimized policies can be tested for genuine faithfulness rather than reward-hacking.

\section{Conclusion}

Verifier-coupled reasoning adds a simple constraint to natural-language rationales: the hidden states over the rationale should encode the programmatically verified claim the model emits. Across synthetic, LeanCheck, KataGo, code, and FEVER settings, this mechanism succeeds as a representation-coupling tool and produces measurable, span-specific verifier alignment. LeanCheck illustrates how rationale-only and proof-only pooling can perfectly separate under counterfactual conflict, and synthetic interventions confirm that consistency-trained activations causally shift claim behavior. The code experiment reveals an equally important finding: coupling reaches 98.6\% by epoch~5 while prose quality never improves, and per-claim analysis shows that fine-grained claims benefit far more from consistency loss than binary claims. FEVER contributes a complementary insight: evidence-only pooling isolates genuine evidence sensitivity at 44\% accuracy, while claim-visible pooling inflates to 99\% by leveraging the claim text rather than parsing evidence. Together, these results show that consistency losses are effective tools for measuring and shaping verifier-aligned representations; closing the remaining gap to fully faithful reasoning is a natural target for verifier-reward RL, architectural information-flow constraints such as the BEC mask, and generation-quality objectives.

\bibliography{verifier_coupled_reasoning_ai4math2026}
\bibliographystyle{icml2026}

\clearpage
\onecolumn
\appendix

\section{Detailed Variant Tables}
\label{app:tables}

\subsection{Synthetic controls and scalar-claim variants}

The synthetic tables show why the auxiliary consistency objective is needed. The language-modeling objective can reach perfect generation while the rationale span carries little verifier information; the consistency-trained variants preserve span-level decodability, counterfactual following, and intervention sensitivity.

\begin{table}[H]
\caption{Synthetic results included in the main narrative. Early small-scale results validate decodability; hard overlapping-vocabulary and intervention runs test robustness and causal effects. In the metric column, cls. means consistency-head classification accuracy, gen. means claim-token generation accuracy, cfact means counterfactual, and orig means original label.}
\centering
\small
\begin{tabular}{llll}
\toprule
Result & Variant & Metric & Value \\
\midrule
Small synthetic & no consistency & cls. acc. & 3.9\% \\
Small synthetic & full sequence & gen. acc. & 93.8\% \\
Small synthetic & rationale only & gen. acc. & 63.0\% \\
Small synthetic & earlier token & gen. acc. & 62.0\% \\
Scaled convergence & all main variants & gen. acc. & 100\% \\
Scaled convergence & no consistency & cls. / cfact cls & 8.8\% / 7.8\% \\
Scaled convergence & no consistency & cfact gen swap & 0\% \\
Negative control & claim-only pooling & cls. acc. & 43\% \\
Negative control & claim-only pooling & cfact swap / orig & 48\% / 8\% \\
Hard overlap & no consistency & cls. / cfact cls & 4.69\% / 6.25\% \\
Hard overlap & no consistency & gen. / cfact gen & 100\% / 100\% \\
Hard overlap & rationale only & gen. / cls. / cfact & 100\% / 100\% / 100\% \\
Hard overlap & full sequence & gen. / cls. / cfact & 81.25\% / 100\% / 100\% \\
Hard overlap & earlier token & gen. / cls. / cfact & 100\% / 100\% / 100\% \\
Hard overlap & shuffled control & consistency variants & 10.16\% \\
Intervention & baseline & layer-0 effect & 31\% \\
Intervention & consistency trained & layer-0 effect & 73--89\% \\
\bottomrule
\end{tabular}
\end{table}

\begin{table}[H]
\caption{Scaled convergence (30 epochs, 5120 train) vs.\ smoke test (5 epochs, 512 train). Key deltas: the no-consistency baseline improves generation from 75\% to 100\% but classifier accuracy stays near chance; consistency-trained variants maintain perfect coupling at convergence. Gen is generation accuracy, Cls is rationale-pooled classifier accuracy, Cfact is counterfactual classifier swap following.}
\label{tab:scaled_vs_smoke}
\centering
\small
\begin{tabular}{lrrrrrr}
\toprule
Variant & Gen$_s$ & Gen$_c$ & Cls$_s$ & Cls$_c$ & Cfact$_s$ & Cfact$_c$ \\
\midrule
\texttt{no\_consistency} & .750 & 1.000 & .039 & .088 & .031 & .078 \\
\texttt{rationale\_only} & .633 & 1.000 & 1.000 & 1.000 & 1.000 & 1.000 \\
\texttt{full\_sequence} & .938 & 1.000 & 1.000 & 1.000 & 1.000 & 1.000 \\
\texttt{earlier\_token} & .617 & 1.000 & 1.000 & 1.000 & 1.000 & 1.000 \\
\bottomrule
\end{tabular}
\end{table}

The generated-rationale scalar experiment is the closest synthetic analogue of the paper's sequence format: a model writes a rationale and emits a scalar or binned claim. This experiment produces the \emph{strongest coupling result} across all settings: both \texttt{rationale\_only} and \texttt{full\_consistency} reach \emph{perfect} 1.000 claim-bin accuracy and perfect counterfactual following (1.000 Orig, 0.000 Swap), meaning the rationale span is a lossless predictor of the verifier-derived bin. This is stronger than LeanCheck (which reaches 100\% but on binary labels) and KataGo (81.1\% on 10-way bins) because the model must generate the rationale tokens before they are pooled, so coupling is achieved through generated text, not just input-conditioned representations. The LM-only baseline reaches 0\% bin accuracy despite 82\% token accuracy, confirming that the standard LM objective does not create claim-predictive rationale states.

\begin{table}[H]
\caption{Generated rationale plus scalar-claim experiment. Token is token-level generation accuracy, Bin acc. is claim-bin accuracy, MSE is mean squared error for the scalar claim, Swap is counterfactual following of the swapped rationale label, and Orig is following of the original label.}
\centering
\small
\begin{tabular}{lrrrrr}
\toprule
Variant & Token & Bin acc. & MSE & Swap & Orig \\
\midrule
\texttt{lm\_only} & .8231 & .0000 & .12835 & .0313 & .1172 \\
\texttt{no\_consistency} & .8197 & .0410 & .00014 & .1602 & .0352 \\
\texttt{rationale\_only} & .8199 & 1.0000 & .09820 & .0000 & 1.0000 \\
\texttt{full\_consistency} & .8166 & 1.0000 & .00018 & .0000 & 1.0000 \\
\texttt{random\_consistency} & .8190 & .1641 & .00029 & .0664 & .1406 \\
\bottomrule
\end{tabular}
\end{table}

\subsection{Shuffled-pairing controls across synthetic variants}

The shuffled-pairing control randomizes rationale--claim assignments at evaluation time: each rationale is paired with a uniformly random claim rather than its correct label. If the consistency head were memorizing surface templates instead of reading rationale content, shuffled accuracy would remain high. \Cref{tab:shuffled} shows that all variants and settings collapse to near the $1/8 = 0.125$ chance level under shuffled pairings, confirming that performance depends on correct rationale--claim alignment.

\begin{table}[H]
\caption{Shuffled-pairing control accuracies across three synthetic settings (8 latent states, chance $= 0.125$). Shuf.\ Gen is generation accuracy under shuffled pairings; Shuf.\ Cls is consistency-head classification accuracy under shuffled pairings. All values are near chance, confirming that the consistency head reads rationale content rather than memorizing patterns.}
\label{tab:shuffled}
\centering
\small
\begin{tabular}{llrr}
\toprule
Setting & Variant & Shuf.\ Gen & Shuf.\ Cls \\
\midrule
\multirow{5}{*}{Smoke test}
 & \texttt{no\_consistency}   & .1016 & .1172 \\
 & \texttt{rationale\_only}   & .0781 & .1016 \\
 & \texttt{full\_sequence}    & .0938 & .1016 \\
 & \texttt{earlier\_token}    & .0859 & .1016 \\
 & \texttt{claim\_only\_pool} & .1016 & .0859 \\
\midrule
\multirow{4}{*}{Scaled run}
 & \texttt{no\_consistency}   & .1289 & .1641 \\
 & \texttt{rationale\_only}   & .1289 & .1289 \\
 & \texttt{full\_sequence}    & .1289 & .1289 \\
 & \texttt{earlier\_token}    & .1289 & .1289 \\
\midrule
\multirow{4}{*}{Hard overlap}
 & \texttt{no\_consistency}   & .1016 & .1719 \\
 & \texttt{rationale\_only}   & .1016 & .1016 \\
 & \texttt{full\_sequence}    & .0938 & .1016 \\
 & \texttt{earlier\_token}    & .1016 & .1016 \\
\bottomrule
\end{tabular}
\end{table}

\subsection{Hidden-state intervention: per-layer, per-variant breakdown}
\label{app:intervention}

The hidden-state intervention experiment provides the strongest causal evidence in the synthetic setting. The methodology is as follows. For each sample pair (an original sequence with latent state $A$ and a swapped sequence with latent state $B$) the evaluation (1) runs a forward pass with the original sequence and caches post-block hidden states at rationale token positions after each transformer block; (2) for each block $i$, runs a second forward pass with the swapped sequence but replaces the hidden states at rationale positions after block $i$ with the cached original states, letting all subsequent blocks process the patched activations; and (3) reads the greedy next-token prediction at the final separator position (immediately before the claim span) to determine whether the model follows the original latent state's claim token or the swapped state's claim token. This procedure directly tests whether rationale-span activations at a given layer causally determine the downstream claim, rather than merely being decodable by a post-hoc probe.

\Cref{tab:intervention} reports the full per-layer, per-variant results. The two-layer model (2 blocks, $d_\text{model} = 64$, 4 heads, $d_\text{ff} = 128$) was trained for 30 epochs on 5{,}120 examples with 8 latent states; intervention was evaluated on 64 sample pairs per variant (seed 42).

\begin{table}[H]
\caption{Hidden-state intervention results by variant and layer. \emph{Follows orig.\ HS} is the fraction of samples where patching rationale hidden states causes the model to predict the original latent state's claim token; \emph{Follows swapped} is the fraction where the model still predicts the swapped sequence's claim token despite the patch. All values are over $n = 64$ sample pairs.}
\label{tab:intervention}
\centering
\small
\begin{tabular}{llrr}
\toprule
Variant & Layer & Follows orig.\ HS & Follows swapped \\
\midrule
\texttt{no\_consistency\_loss} & 0 & .3125 & .6875 \\
\texttt{no\_consistency\_loss} & 1 & .0000 & 1.0000 \\
\midrule
\texttt{rationale\_only}      & 0 & .8906 & .1094 \\
\texttt{rationale\_only}      & 1 & .0000 & 1.0000 \\
\midrule
\texttt{full\_sequence}       & 0 & .7500 & .2500 \\
\texttt{full\_sequence}       & 1 & .0000 & 1.0000 \\
\midrule
\texttt{earlier\_token\_only} & 0 & .7344 & .2500 \\
\texttt{earlier\_token\_only} & 1 & .0000 & 1.0000 \\
\bottomrule
\end{tabular}
\end{table}

Two patterns stand out. First, the \textbf{layer-0 vs.\ layer-1 contrast} is sharp. At layer~0 (after the first transformer block), consistency-trained variants show 73--89\% of samples following the patched original hidden states, compared with only 31\% for the no-consistency baseline. At layer~1 (after the final block), \emph{every} variant, including all consistency-trained models, shows 0\% following the original hidden states and 100\% following the swapped tokens. This is expected: patching at the last layer is too late because the model has already committed to a prediction based on the full forward pass through both blocks. The intervention can only redirect behavior when it replaces activations early enough for downstream computation to propagate the patched signal.

Second, the \texttt{earlier\_token\_only} variant reveals a subtle \textbf{third-state phenomenon}. At layer~0, 73.4\% of samples follow the original hidden states and 25.0\% follow the swapped tokens, but these do not sum to 100\%: the remaining $\sim$1.6\% of samples predict a token that matches neither the original nor the swapped claim. In these cases the patched activations push the model into a third latent state not present in either input, suggesting that the intervention creates an out-of-distribution activation pattern for a small fraction of samples. The other consistency-trained variants (\texttt{rationale\_only} and \texttt{full\_sequence}) sum to 100\%, indicating clean binary switching behavior under intervention.

\subsection{LeanCheck formal-verifier results}

LeanCheck is the cleanest formal-verifier result because the labels are mechanically generated by Lean and the counterfactuals separate proof and rationale evidence. The training set contains 1,000 examples distributed roughly uniformly across six domains: addition commutativity (182), list append-nil (179), conjunction introduction (175), reflexive equality (158), propositional hypothesis (157), and addition right-identity (149), with a balanced label split (520 VERIFIES, 480 FAILS). Among the 480 negative examples, the most common mutation types are replace-tactic (107), delete-final-line (102), and rename-variable (95), followed by wrong-lemma (63), wrong-pairing (62), adversarial-near-miss (59), and missing-premise (12). The 100 minimal pairs use primarily wrong-lemma (83) and rename-variable (17) mutations, chosen for their single-line minimal difference.

\begin{table}[H]
\caption{LeanCheck GPT-2 run with Lean-derived labels. Cls is consistency-head classification accuracy, Gen is claim-token generation accuracy, Swap is counterfactual following of the swapped rationale label, Proof is following of the proof label, and Pair is minimal-pair flip accuracy.}
\label{tab:leancheck}
\centering
\small
\begin{tabular}{lrrrrr}
\toprule
Variant & Cls & Gen & Swap & Proof & Pair \\
\midrule
\texttt{lm\_only} & .490 & .510 & .495 & .495 & .000 \\
\texttt{no\_consistency} & .510 & .525 & .505 & .505 & .000 \\
\texttt{rationale\_only} & 1.000 & .640 & 1.000 & .010 & 1.000 \\
\texttt{full\_sequence} & 1.000 & .990 & .995 & .005 & 1.000 \\
\texttt{proof\_only} & 1.000 & .915 & .010 & 1.000 & 1.000 \\
\texttt{random\_consistency} & .410 & .775 & .400 & .590 & .070 \\
\texttt{wrong\_span} & .520 & .540 & .495 & .515 & .000 \\
\bottomrule
\end{tabular}
\end{table}

The activation-patching table shows a clear pattern consistent with the diagnostic ladder. Head-logit effects are large for the intended rationale-pooled model and close to zero or negative for several controls, which is the pattern predicted by the diagnostic ladder.

\begin{table}[H]
\caption{LeanCheck activation patching summary. LM denotes language-model claim-token logits, Head denotes consistency-head logits, rat denotes rationale patching, and rnd denotes random-token patching. Head patching is more discriminative than LM-logit patching.}
\centering
\small
\begin{tabular}{lrrrr}
\toprule
Variant & LM rat & LM rat-rnd & Head rat & Head rat-rnd \\
\midrule
\texttt{lm\_only} & 6.086 & 5.410 & .186 & -.001 \\
\texttt{no\_consistency} & 3.718 & 5.097 & .001 & .064 \\
\texttt{rationale\_only} & 5.254 & 9.774 & 11.584 & 4.518 \\
\texttt{full\_sequence} & 9.152 & 5.463 & 7.867 & .224 \\
\texttt{proof\_only} & 1.410 & -5.208 & .000 & -3.758 \\
\texttt{random\_consistency} & 6.135 & -7.494 & -.054 & .025 \\
\texttt{wrong\_span} & 9.574 & 10.741 & .000 & -.014 \\
\bottomrule
\end{tabular}
\end{table}

\subsection{KataGo and Go claim-head results}

KataGo tests whether the mechanism survives a dense domain verifier rather than a binary proof checker. The key pattern is that consistency-trained commentary spans encode engine-derived win-rate buckets, while LM-only, no-consistency, and random-label controls remain much weaker.

\begin{table}[H]
\caption{KataGo win-probability coupling. Bin acc. is claim-bin accuracy from commentary-span states, MSE and MAE are scalar win-rate errors, Pearson and Spearman are win-rate correlations, Swap is counterfactual following of the swapped commentary label, and Orig is following of the original label.}
\centering
\small
\begin{tabular}{lrrrrrrr}
\toprule
Variant & Bin acc. & MSE & MAE & Pearson & Spearman & Swap & Orig \\
\midrule
\texttt{lm\_only} & .016 & .1606 & .3785 & .362 & .487 & .10 & .11 \\
\texttt{no\_consistency} & .251 & .0016 & .0268 & .995 & .932 & .20 & .21 \\
\texttt{rationale\_only} & .787 & .1807 & .3962 & .154 & .188 & .52 & .14 \\
\texttt{full\_consistency} & .811 & .0020 & .0280 & .994 & .918 & .55 & .14 \\
\texttt{random\_consistency} & .160 & .0017 & .0284 & .995 & .935 & .17 & .10 \\
\bottomrule
\end{tabular}
\end{table}

The Go GPT-OSS tables summarize a supervised claim-head experiment on Go positions. A GPT-OSS model was trained on a 1k train / 200 evaluation split to predict multiple structured claims derived from board state and KataGo-style analysis targets, including win-probability bins, score-lead bins, phase, contested regions, move urgency, and search surprise. This is not the same as the KataGo win-probability consistency experiment above: it tests whether a Go-language model can learn a broader multi-claim ontology before adding stronger verifier-coupling or RL objectives. Aggregate accuracy rises with training, but macro-F1 reveals class collapse; the model learns frequent labels much faster than rare or fine-grained tactical categories. A zero-shot baseline using SmolLM-3B (without fine-tuning) produces largely degenerate outputs on the same positions: repetitive prompt echoing, role-tag cycling, and no substantive Go commentary, confirming that domain-specific fine-tuning is necessary for coherent position analysis.

\begin{table}[H]
\caption{Go GPT-OSS aggregate claim-head training on the 1k/200 split. Acc. denotes accuracy and F1 denotes macro-F1 averaged across label classes. Macro-F1 peaks before raw accuracy, indicating overfitting or majority-class collapse after epoch 4.}
\centering
\small
\begin{tabular}{lrrrr}
\toprule
Epoch & Train acc. & Eval acc. & Train F1 & Eval F1 \\
\midrule
1 & .4858 & .4889 & .1953 & .1955 \\
2 & .5272 & .5228 & .2440 & .2375 \\
3 & .5561 & .5539 & .2747 & .2674 \\
4 & .5873 & .5544 & .3964 & .3206 \\
5 & .6288 & .5617 & .4573 & .3123 \\
\bottomrule
\end{tabular}
\end{table}

\begin{table}[H]
\caption{Go GPT-OSS per-claim results at epoch 4. Accuracy is raw per-head label accuracy and Macro-F1 is class-balanced F1. Global contestedness is the best-behaved head; several other heads have acceptable raw accuracy but weak macro-F1 because predictions collapse toward frequent labels.}
\centering
\small
\begin{tabular}{lrr}
\toprule
Claim head & Accuracy & Macro-F1 \\
\midrule
\texttt{win\_prob\_bin} & .315 & .177 \\
\texttt{score\_lead\_bin} & .550 & .171 \\
\texttt{phase\_estimate} & .755 & .326 \\
\texttt{main\_control\_region} & .300 & .191 \\
\texttt{main\_contested\_region} & .450 & .342 \\
\texttt{global\_contestedness} & .865 & .765 \\
\texttt{best\_move\_region} & .550 & .276 \\
\texttt{move\_urgency} & .540 & .324 \\
\texttt{search\_surprise} & .665 & .313 \\
\bottomrule
\end{tabular}
\end{table}

\subsection{Code coupling and failure-mode results}

Hidden-state coupling is high in several code variants, but manual review shows the generated explanations are not usable. Thus, probe or swap success cannot be equated with faithful explanation.

\begin{table}[H]
\caption{Code coupling rich-ontology results. Coupling is mean per-claim classifier accuracy from the selected explanation span; Swap is counterfactual swap influence. Generation-quality metrics for this experiment are reported separately in \Cref{tab:code_bpe} (a subword-tokenizer re-run); BLEU/ROUGE from the original word-level decode were artifacts and are omitted here.}
\label{tab:code_rich}
\centering
\small
\begin{tabular}{lrr}
\toprule
Variant & Coupling & Swap \\
\midrule
\texttt{consistency\_loss} & .986 & .978 \\
\texttt{no\_consistency} & .286 & $-.033$ \\
\texttt{claim\_only} & .595 & .122 \\
\texttt{random\_label} & .562 & $-.022$ \\
\texttt{no\_claim\_attn} & .986 & .978 \\
\texttt{claims\_from\_expl} & .986 & .978 \\
\bottomrule
\end{tabular}
\end{table}

The simpler V1 and strict-flow variants show the same pattern from another angle: coupling can be made almost perfect with consistency loss and masking, but surface bottlenecks are harder and the resulting prose remains poor.

\begin{table}[H]
\caption{Code coupling variants beyond the rich-ontology table. Coupling is mean per-claim classifier accuracy, Swap is counterfactual swap influence, Claim acc. is emitted structured-claim accuracy, and BLEU-1 is unigram overlap. V1 establishes the basic effect; V2 shows that strict hidden-state flow can preserve coupling, while surface bottlenecks weaken or invert swap behavior.}
\centering
\small
\begin{tabular}{lrrrr}
\toprule
Variant & Coupling & Swap & Claim acc. & BLEU-1 \\
\midrule
V1 \texttt{consistency\_loss} & 1.000 & 1.000 & 1.000 & .067 \\
V1 \texttt{no\_consistency} & .175 & .050 & -- & .060 \\
V2 \texttt{no\_claim\_attn} & 1.000 & 1.000 & 1.000 & .076 \\
V2 \texttt{claims\_from\_expl} & 1.000 & 1.000 & .800 & .071 \\
V2 \texttt{surface\_bottleneck} & .697 & -.150 & 1.000 & .068 \\
V2 \texttt{surface\_no\_expl\_lm} & .808 & -.050 & .100 & .003 \\
\bottomrule
\end{tabular}
\end{table}

\begin{table}[H]
\caption{Manual qualitative review of code explanations. Behavior is behavior-correct prose, Full prose is fully correct prose, and Time/Space/Bug are manually checked claim-value correctness counts out of 20. All variants fail the fully-correct prose and behavior-correct criteria, even when structured claims are often individually correct. Prose is fluent under a standard subword tokenizer (\Cref{tab:code_bpe}); the failure is infidelity, not incoherence. The Time/Space/Bug counts reflect the word-level run and understate structured-claim correctness.}
\centering
\small
\begin{tabular}{lrrrrr}
\toprule
Variant & Behavior & Full prose & Time & Space & Bug \\
\midrule
\texttt{consistency\_loss} & 0/20 & 0/20 & 4/20 & 10/20 & 17/20 \\
\texttt{claim\_only} & 0/20 & 0/20 & 5/20 & 16/20 & 17/20 \\
\texttt{no\_consistency} & 0/20 & 0/20 & 5/20 & 11/20 & 17/20 \\
\texttt{random\_label} & 0/20 & 0/20 & 5/20 & 9/20 & 17/20 \\
\bottomrule
\end{tabular}
\end{table}

\subsection{Code coupling claim ontology and architectural variants}

The rich-ontology code experiment uses 12 structured claim heads, each with an oracle-derived label: \texttt{time\_complexity} (6 values: $O(1)$ through $O(n^3)$), \texttt{space\_complexity} (4), \texttt{best\_case\_time} (6), \texttt{algorithm\_class} (7: brute-force, divide-and-conquer, greedy, DP, etc.), \texttt{loop\_structure} (6), \texttt{key\_operation} (7: comparison, arithmetic, hashing, etc.), \texttt{access\_pattern} (5), \texttt{auxiliary\_structures} (7), \texttt{mutates\_input} (2), \texttt{correctness\_status} (5), \texttt{handles\_empty\_input} (3), and \texttt{handles\_duplicates} (4)---62 total claim-value tokens. Per-claim difficulty varies: binary claims like \texttt{mutates\_input} and \texttt{handles\_empty\_input} reach near-perfect accuracy even without consistency loss, while fine-grained claims like \texttt{algorithm\_class} lag at 13.4\% without consistency loss versus 98.6\% with it.

The V2 code experiments test four architectural variants beyond the standard consistency-loss ablation. \texttt{no\_claim\_to\_claim\_attention} adds an additive attention bias that blocks claim tokens from attending to other claim tokens, forcing each claim to be predicted independently from the explanation span. \texttt{claims\_from\_explanation\_only} restricts claim tokens to attend only to explanation tokens, enforcing a code$\to$explanation$\to$claim information path. Both achieve coupling near 1.0 and swap influence near 1.0, matching the standard consistency-loss variant. \texttt{surface\_bottleneck\_consistency} replaces hidden-state pooling with classification over softmax probability distributions at explanation positions, testing whether coupling can survive a surface-level bottleneck. This variant drops to 0.697 coupling and $-0.15$ swap influence, showing that the consistency signal requires access to hidden states, not just output distributions. \texttt{surface\_no\_expl\_lm} further disables LM loss on explanation tokens, collapsing to BLEU 0.003 by epoch 3; the model produces no readable text at all.

Beyond selection sort (\Cref{fig:code-ex}), the same pattern appears across other functions under the subword-tokenizer re-run (\Cref{app:code_bpe}). For \texttt{caesar\_cipher}, the model emits correct structured claims but generates fluent prose describing a different algorithm: ``\texttt{longest\_palindrome\_substr} expands around every center (odd and even length) to find the longest palindromic substring. However, the final line uses stale \texttt{l} and \texttt{r} from the last expansion, which is a \texttt{wrong\_condition} bug. Time $O(n^2)$; space $O(1)$ \ldots'' For \texttt{is\_prime}, the prose says ``\texttt{binary\_search\_rec} implements binary search recursively, halving the search range each call and reaching $O(\log n)$ time in the worst case. The recursion stack uses $O(\log n)$ space \ldots'' In every case the prose is fluent and describes an unrelated algorithm while the 12 structured claim tags are emitted correctly. The earlier ``garbled'' word-level decodes (e.g.\ ``Computesprefixsumarray.\ O(n) time and O(n) space'') were tokenizer artifacts rather than generation failures; the substantive failure is infidelity, not incoherence.

\subsection{Code generation under a subword tokenizer}
\label{app:code_bpe}

The code-experiment generation metrics in the tables above were decoded with a custom word-level tokenizer. To check whether the apparent generation failure was a property of the model or of that decoder, we re-ran the 12-claim rich ontology with a standard GPT-2 BPE tokenizer, on a pretrained and a from-scratch GPT-2 (124M) of identical architecture (19 templates, 3{,}000 examples, 20 epochs, seed~42). \Cref{tab:code_bpe} reports the result. Representation coupling is unchanged: the consistency objective drives all 12 claim heads to perfect accuracy, reproducing the $\approx$0.986 of the word-level run. The generation-quality picture, however, is entirely different from the word-level decode: BLEU-1 is $\approx$0.29 rather than 0.06, the 4-gram repetition rate is 0.000 (no degeneration), and the structured claim tags are emitted with $\approx$98\% accuracy rather than 2.2\%. Manual inspection confirms the prose is fluent, grammatical, multi-sentence English. The earlier ``garbled,'' ``separator-token repetition,'' and ``2.2\% claim emission'' observations were artifacts of the word-level tokenizer's decoder, not generation failures.

The substantive failure survives the tokenizer change: the fluent prose is \emph{unfaithful}, describing a different algorithm than the input code (for example, \texttt{caesar\_cipher} is explained as \texttt{longest\_palindrome\_substr}, and \texttt{is\_prime} as \texttt{binary\_search}). Pretrained and from-scratch models behave identically. The 12-claim ontology thus tells the same story as the main-text code experiment: near-perfect coupling, fluent generation, unfaithful content. Scope: the BPE re-run covers \texttt{consistency\_loss}, \texttt{no\_consistency}, and \texttt{random\_label} on 19 reconstructed templates; the \texttt{claim\_only} and V2 attention variants were not re-run. Their Coupling/Swap numbers (tokenizer-independent) are retained in \Cref{tab:code_rich}, and their original generation-quality numbers should be treated as superseded by the subword-tokenizer evidence below.

\begin{table}[H]
\caption{Code coupling, 12-claim rich ontology, re-run with a standard GPT-2 BPE tokenizer. Coupling and Swap reproduce the word-level run; BLEU-1, ROUGE-L, repetition, and claim emission supersede the word-level figures, which were decoder artifacts. Pretrained and from-scratch GPT-2 are equivalent.}
\label{tab:code_bpe}
\centering
\small
\begin{tabular}{llrrrrrr}
\toprule
Condition & Variant & Coupling & Swap & BLEU-1 & ROUGE-L & Repet. & Claim emit \\
\midrule
Pretrained   & \texttt{consistency\_loss}   & 1.000 & 1.000 & .294 & .189 & .000 & .983 \\
Pretrained   & \texttt{no\_consistency}     & .185 & $-.002$ & .296 & .190 & .000 & .983 \\
Pretrained   & \texttt{random\_label}       & .550 & $-.021$ & .271 & .180 & .000 & 1.000 \\
From scratch & \texttt{consistency\_loss}   & 1.000 & 1.000 & .286 & .173 & .000 & .983 \\
From scratch & \texttt{no\_consistency}     & .317 & $-.040$ & .407 & .302 & .000 & .975 \\
From scratch & \texttt{random\_label}       & .544 & $-.021$ & .407 & .302 & .000 & .975 \\
\bottomrule
\end{tabular}
\end{table}

\begin{table}[H]
\caption{Training dynamics of the BPE re-run (pretrained GPT-2, 12-claim ontology). Coupling reaches 1.000 by epoch~2 and claim emission $\approx$100\% by epoch~5; the repetition rate is 0.000 from epoch~2 onward and BLEU-1 stays flat near 0.3 (fluent throughout). Faithful generation never emerges at any epoch.}
\label{tab:code_bpe_dynamics}
\centering
\small
\begin{tabular}{llrrrrr}
\toprule
Variant & Epoch & Coupling & Swap & BLEU-1 & Claim emit & Repet. \\
\midrule
\texttt{consistency}    & 1  & .873  & .778  & .345 & .517  & .203 \\
\texttt{consistency}    & 5  & 1.000 & 1.000 & .275 & 1.000 & .000 \\
\texttt{consistency}    & 10 & 1.000 & 1.000 & .321 & 1.000 & .000 \\
\texttt{consistency}    & 20 & 1.000 & 1.000 & .294 & .983  & .000 \\
\addlinespace
\texttt{no\_consistency} & 1  & .268 & $-.033$ & .309 & .625  & .000 \\
\texttt{no\_consistency} & 5  & .183 & $-.004$ & .276 & 1.000 & .000 \\
\texttt{no\_consistency} & 10 & .206 & .003    & .322 & 1.000 & .000 \\
\texttt{no\_consistency} & 20 & .185 & $-.002$ & .296 & .983  & .000 \\
\bottomrule
\end{tabular}
\end{table}

\subsection{Capacity and loss-weight controls for the code experiment}
\label{app:capacity_lambda}

The two controlled experiments referenced in \Cref{sec:ceilings} (capacity control and loss-weight sweep) are reported in full below. They isolate the decodability--faithfulness gap from two natural confounds: a capacity-starved from-scratch model, and a consistency loss that may have been under-weighted relative to the language-modeling loss.

\begin{table}[H]
\caption{Capacity control: the code-coupling consistency objective on a \emph{pretrained} vs.\ a \emph{from-scratch} GPT-2 (124M) of identical architecture and subword tokenizer (mean over 3 seeds). Coupling responds to the consistency loss and is identical across conditions; generated-prose quality (BLEU-1 vs.\ the reference explanation) and degeneration (4-gram repetition rate) are unaffected by pretraining. The decodability--faithfulness gap does not narrow with real generative capacity.}
\label{tab:capacity}
\centering
\small
\begin{tabular}{llrrr}
\toprule
Condition & Variant & Coupling & BLEU-1 & Repetition \\
\midrule
Pretrained   & \texttt{consistency\_loss}    & 1.000 & .191 & .000 \\
Pretrained   & \texttt{no\_consistency\_loss} & .393 & .182 & .000 \\
From scratch & \texttt{consistency\_loss}    & 1.000 & .183 & .000 \\
From scratch & \texttt{no\_consistency\_loss} & .339 & .178 & .000 \\
\bottomrule
\end{tabular}
\end{table}

\begin{table}[H]
\caption{Loss-weight sweep: the \texttt{consistency\_loss} variant on pretrained GPT-2 with $\lambda_{\mathrm{cons}}$ varied over four orders of magnitude (20 epochs, seed 42). Coupling is pinned at $1.00$ and every generation metric is flat. The consistency loss saturates within the first epoch, so its gradient, and hence $\lambda_{\mathrm{cons}}$'s influence, vanishes; the final LM loss is identical across the sweep. Reweighting has no lever on generation.}
\label{tab:lambda}
\centering
\small
\begin{tabular}{rrrrrr}
\toprule
$\lambda_{\mathrm{cons}}$ & Coupling & BLEU-1 & Prose acc. & LM loss & Cons.\ loss \\
\midrule
$1$      & 1.000 & .201 & .30 & .0624 & 3.9e\!-\!5 \\
$5$      & 1.000 & .205 & .25 & .0623 & 7e\!-\!6 \\
$25$     & 1.000 & .207 & .25 & .0623 & 1e\!-\!6 \\
$100$    & 1.000 & .215 & .25 & .0625 & $\approx$0 \\
$1000$   & 1.000 & .219 & .30 & .0622 & $\approx$0 \\
$10000$  & 1.000 & .200 & .20 & .0631 & $\approx$0 \\
\bottomrule
\end{tabular}
\end{table}

\subsection{Per-claim difficulty breakdown}
\label{tab:perclaim}

The per-claim analysis for the 12-claim code ontology reveals that consistency loss has differential impact across claim types. Binary and low-cardinality claims are partially learnable from code structure alone; fine-grained categorical claims are where the consistency objective provides the largest gain.

\begin{table}[H]
\caption{Per-claim accuracy at epoch~20 for \texttt{consistency\_loss} vs.\ \texttt{no\_consistency\_loss} in the 12-claim code ontology. $|\mathcal{Y}|$ is the number of label classes. $\Delta$ is the accuracy gain from consistency training. Binary claims (\texttt{mutates\_input}, \texttt{handles\_empty\_input}) have small $\Delta$ because code structure alone is locally predictive; fine-grained claims show large $\Delta$.}
\centering
\small
\begin{tabular}{lrrrr}
\toprule
Claim head & $|\mathcal{Y}|$ & Cons. & No cons. & $\Delta$ \\
\midrule
\texttt{time\_complexity} & 6 & .986 & .198 & +.788 \\
\texttt{space\_complexity} & 4 & .982 & .416 & +.566 \\
\texttt{best\_case\_time} & 6 & .984 & .106 & +.878 \\
\texttt{algorithm\_class} & 7 & .990 & .134 & +.856 \\
\texttt{loop\_structure} & 6 & .976 & .256 & +.720 \\
\texttt{key\_operation} & 7 & .998 & .096 & +.902 \\
\texttt{access\_pattern} & 5 & .978 & .092 & +.886 \\
\texttt{auxiliary\_structures} & 7 & .972 & .216 & +.756 \\
\texttt{mutates\_input} & 2 & .988 & .774 & +.214 \\
\texttt{correctness\_status} & 5 & .992 & .388 & +.604 \\
\texttt{handles\_empty\_input} & 3 & 1.000 & .530 & +.470 \\
\texttt{handles\_duplicates} & 4 & .986 & .230 & +.756 \\
\midrule
Mean & -- & .986 & .286 & +.700 \\
\bottomrule
\end{tabular}
\end{table}

\subsection{Training dynamics: coupling emerges early, prose quality does not}

The epoch-by-epoch metrics for the code experiment show that coupling strength rises to near-ceiling within the first five epochs, while explanation quality (BLEU-1) never improves meaningfully for any variant. This asymmetry illustrates why consistency losses alone do not close the gap to faithful generation.

\begin{table}[H]
\caption{Code coupling training dynamics (12-claim rich ontology). Coupling strength reaches 98.6\% by epoch~5 and stays there; the no-consistency baseline never exceeds 29\%. Generation-quality dynamics for this setting are reported in \Cref{tab:code_bpe_dynamics} (subword-tokenizer re-run); the BLEU-1 and claim-accuracy columns of the original word-level run were decoder artifacts and are omitted.}
\label{tab:code_dynamics}
\centering
\small
\begin{tabular}{llrr}
\toprule
Variant & Epoch & Coupling & Swap \\
\midrule
\texttt{consistency} & 1 & .840 & .539 \\
\texttt{consistency} & 5 & .986 & .978 \\
\texttt{consistency} & 10 & .986 & .978 \\
\texttt{consistency} & 20 & .986 & .978 \\
\addlinespace
\texttt{no\_consistency} & 1 & .144 & .050 \\
\texttt{no\_consistency} & 5 & .231 & .000 \\
\texttt{no\_consistency} & 10 & .185 & $-.050$ \\
\texttt{no\_consistency} & 20 & .286 & $-.033$ \\
\bottomrule
\end{tabular}
\end{table}

\begin{figure}[H]
\centering
\includegraphics[width=0.48\textwidth]{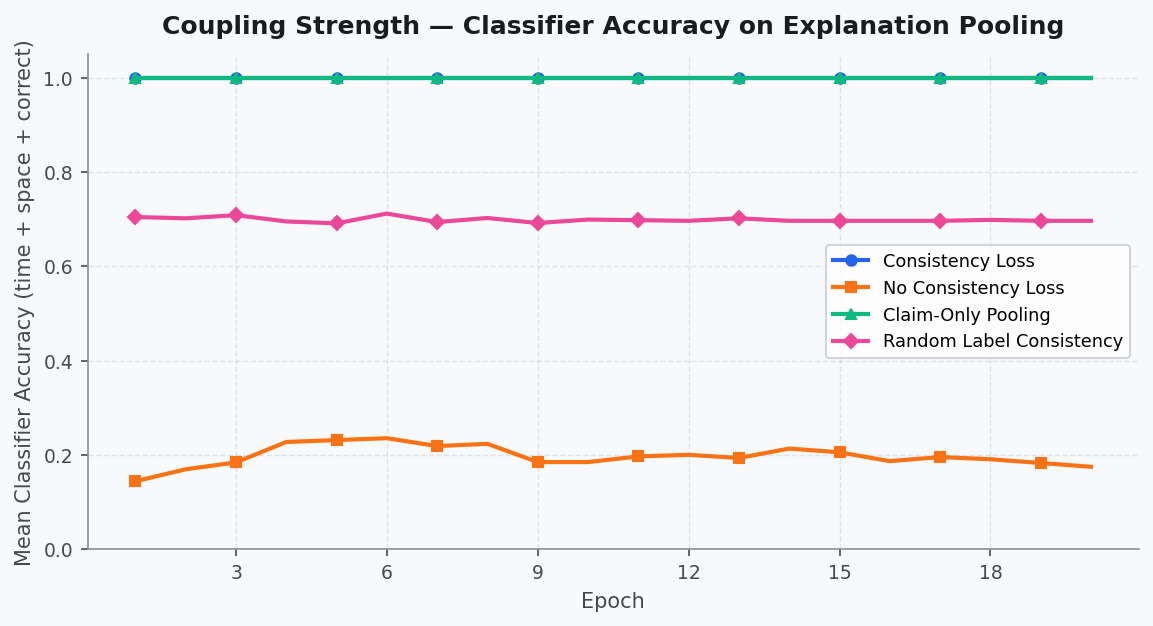}\hfill
\includegraphics[width=0.48\textwidth]{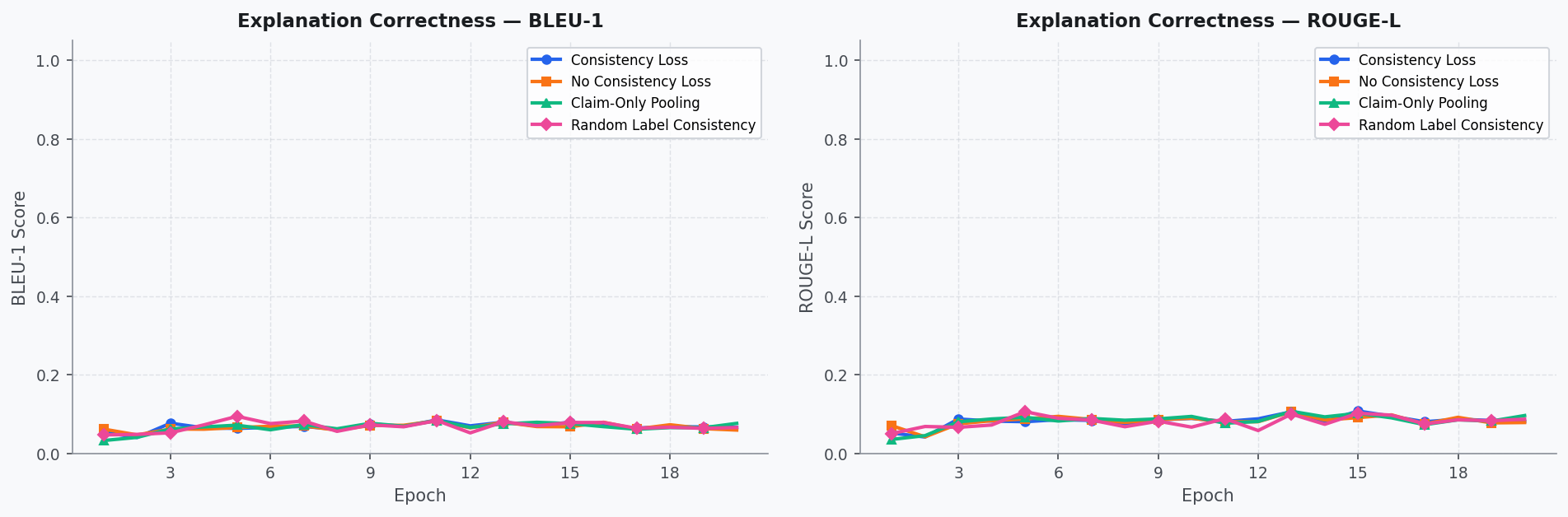}

\medskip
\includegraphics[width=0.48\textwidth]{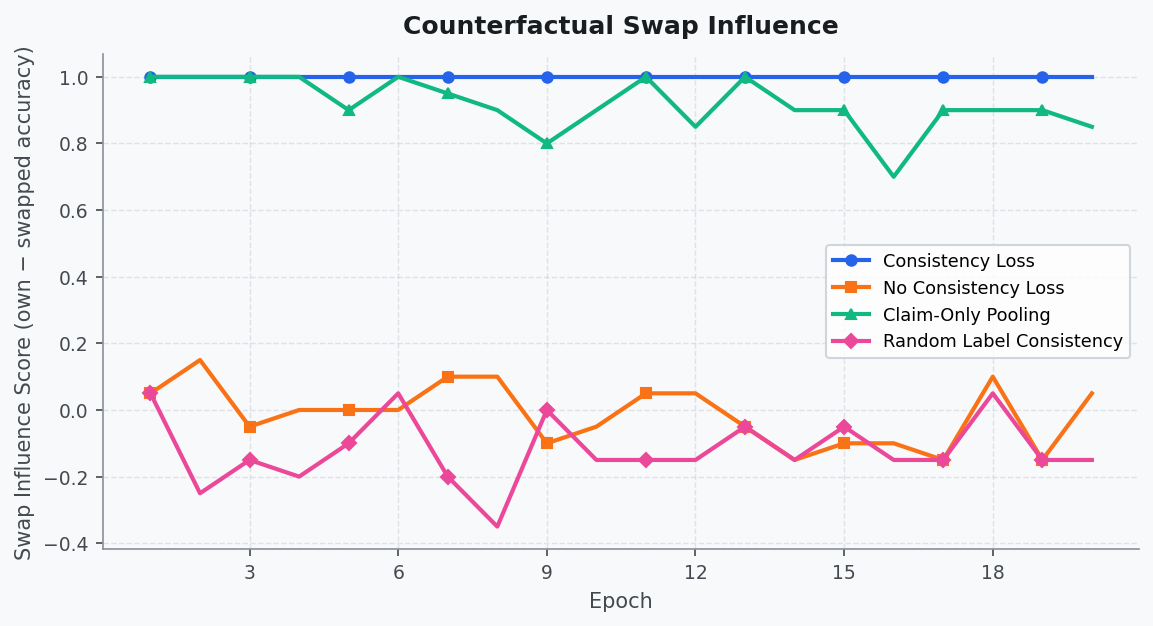}
\caption{\textbf{Code coupling training dynamics (V1, 20 epochs).} \emph{Top left:} Coupling strength (classifier accuracy from explanation-span pooling) reaches 1.0 by epoch~2 for the consistency-loss variant and stays there; baselines remain below 0.25. \emph{Top right:} BLEU-1 and ROUGE-L stay flat near 0.06--0.10 for all variants, prose quality is never learned. \emph{Bottom:} Counterfactual swap influence is 1.0 from epoch~1 for consistency loss; baselines fluctuate near zero. Together these panels show that coupling is learned immediately while explanation quality is never learned, supporting the SFT ceiling argument.}
\label{fig:code_dynamics}
\end{figure}

\Cref{fig:code_dynamics} visualizes this asymmetry. The coupling-strength plot shows that the consistency-loss variant reaches perfect classifier accuracy from the explanation span by epoch~2 and remains there; the no-consistency baseline never exceeds 0.25. In contrast, the explanation-correctness panel shows that BLEU-1 stays flat near 0.06 for \emph{all} variants across all 20 epochs, so consistency loss does not improve prose quality at all. The counterfactual swap plot confirms that consistency-trained representations reliably follow the swapped explanation (swap influence $= 1.0$ from epoch~1), while baselines fluctuate near zero.

\subsection{V2 manual review: strict attention variants}
\label{tab:v2_manual}

The V2 code variants add architectural constraints (no-claim-to-claim attention, claims-from-explanation-only, surface bottleneck). Manual review of 20~examples per variant confirms that none produce usable prose, even when structured claim accuracy is perfect.

\begin{table}[H]
\caption{V2 manual review summary. Behavior is behavior-correct prose, Full prose is fully correct prose, Claims correct counts structured claim correctness. Val coupling is validation coupling strength. All V2 variants fail on prose quality; \texttt{no\_claim\_attn} has perfect structured claims.}
\centering
\small
\begin{tabular}{lrrrr}
\toprule
Variant & Behavior & Full prose & Claims & Coupling \\
\midrule
\texttt{no\_claim\_attn} & 0/20 & 0/20 & 20/20 & 1.000 \\
\texttt{claims\_from\_expl} & 0/20 & 0/20 & 7/20 & 1.000 \\
\texttt{surface\_bottleneck} & 0/20 & 0/20 & 20/20 & 0.697 \\
\texttt{surface\_no\_expl\_lm} & 0/20 & 0/20 & 0/20 & 0.808 \\
\bottomrule
\end{tabular}
\end{table}

\begin{figure}[H]
\centering
\includegraphics[width=0.48\textwidth]{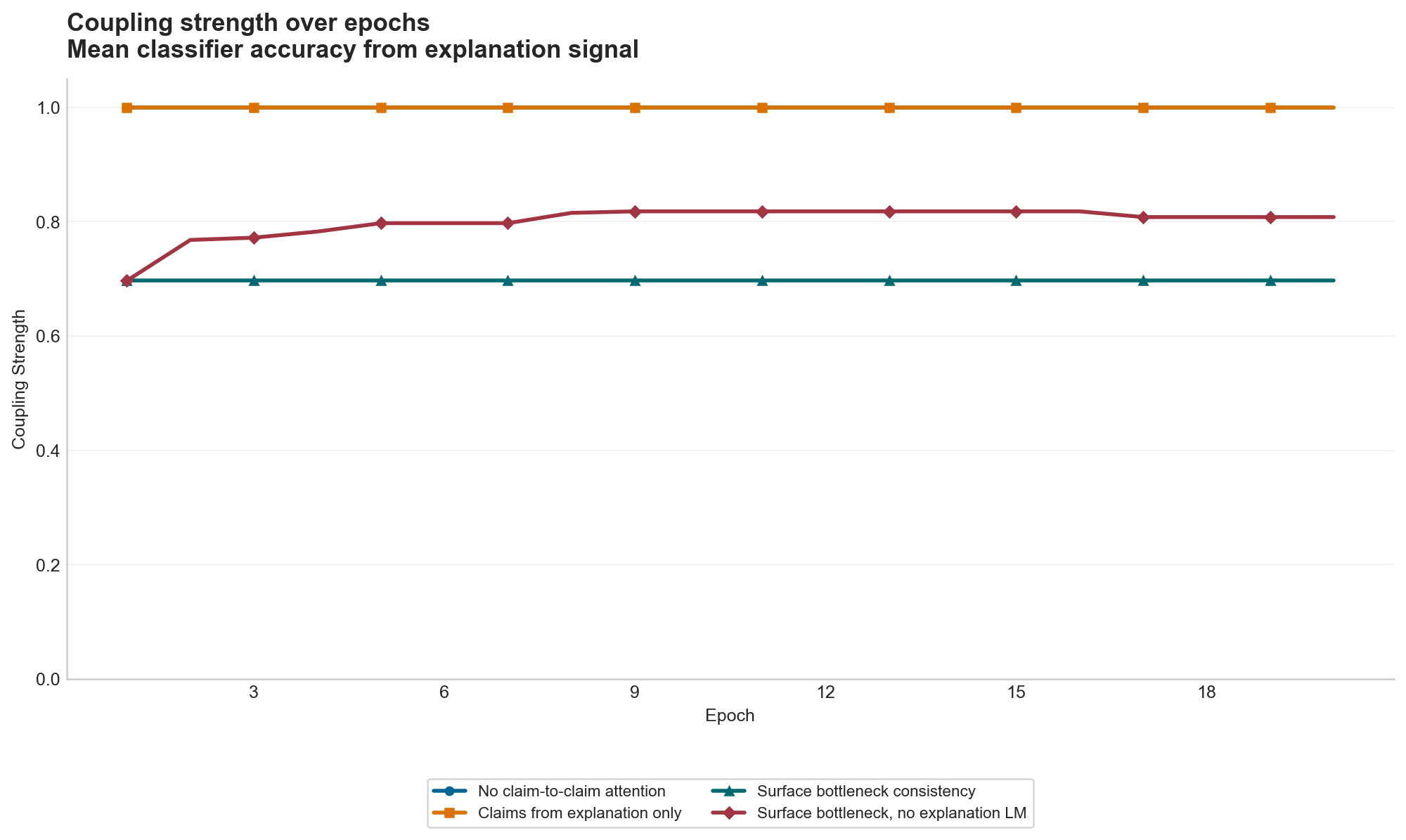}\hfill
\includegraphics[width=0.48\textwidth]{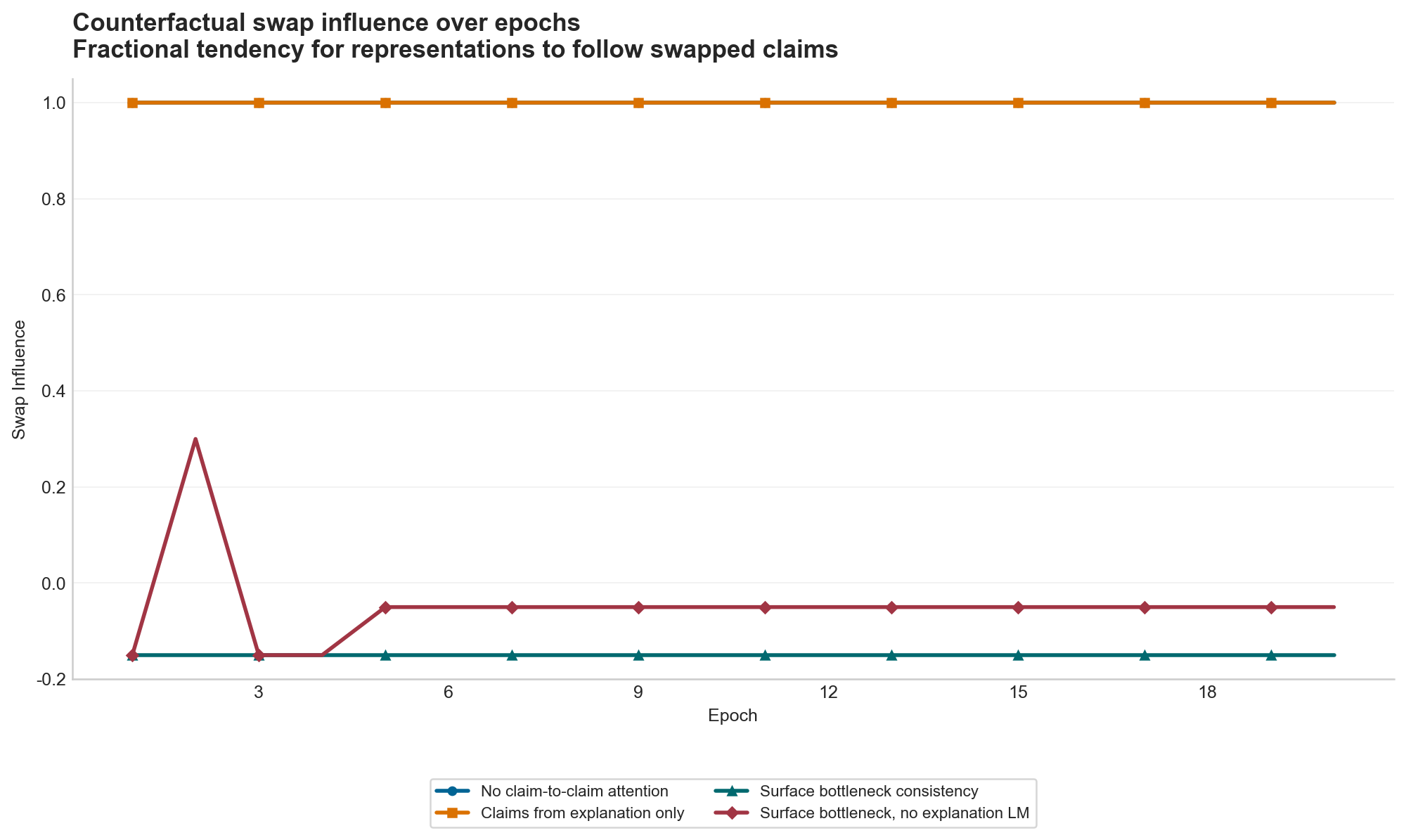}
\caption{\textbf{V2 architectural-variant training dynamics (20 epochs).} \emph{Left:} Coupling strength over training for four V2 variants. \texttt{No claim-to-claim attention} and \texttt{claims from explanation only} maintain 1.0 coupling throughout; surface-bottleneck variants plateau at 0.70--0.81. \emph{Right:} Counterfactual swap influence. Hidden-state variants follow swapped explanations (1.0); surface-bottleneck variants do not ($\sim -0.15$). This confirms that the consistency signal propagates through hidden states, not output distributions.}
\label{fig:code_dynamics_v2}
\end{figure}

\Cref{fig:code_dynamics_v2} shows the V2 architectural-variant training dynamics. The \texttt{no\_claim\_to\_claim\_attention} and \texttt{claims\_from\_explanation\_only} variants maintain perfect coupling strength (1.0) from epoch~1, identical to the standard consistency-loss variant, while the surface-bottleneck variants plateau at 0.70--0.81, confirming that consistency coupling requires access to hidden states rather than softmax probabilities. The counterfactual swap panel mirrors this pattern: strict information-flow variants show swap influence of 1.0 throughout, while surface bottleneck variants hover near $-0.15$, indicating the model does not follow swapped explanations when restricted to surface-level statistics.

\begin{figure}[H]
\centering
\includegraphics[width=0.48\textwidth]{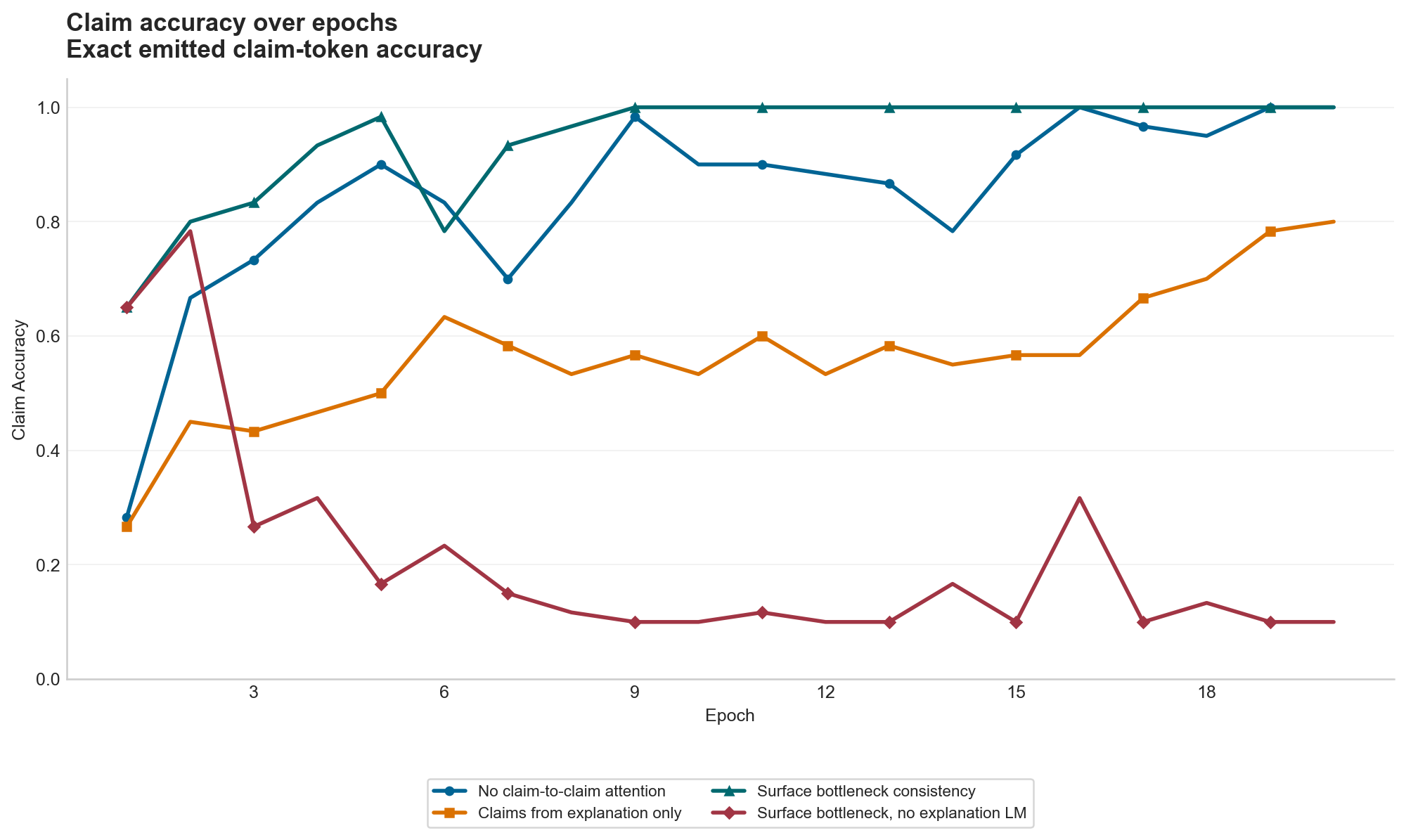}\hfill
\includegraphics[width=0.48\textwidth]{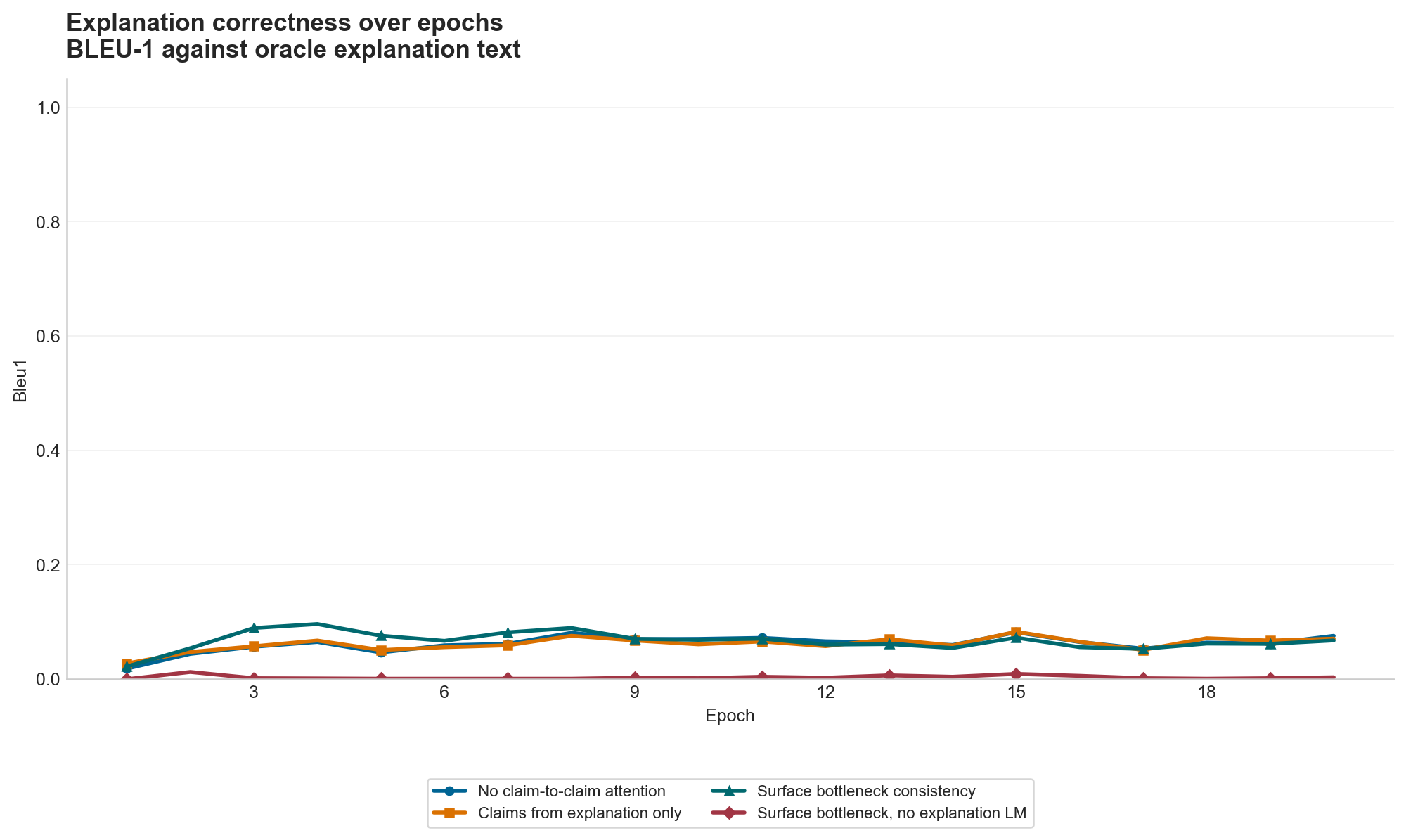}
\caption{\textbf{V2 claim accuracy and explanation correctness (20 epochs).} \emph{Left:} Claim-emission accuracy diverges across architectural variants: \texttt{no\_claim\_attn} and \texttt{surface\_bottleneck} reach $\sim$1.0; \texttt{claims\_from\_expl} plateaus at $\sim$0.80; \texttt{surface\_no\_expl\_lm} collapses below 0.15. \emph{Right:} BLEU-1 stays flat near 0.06 for all variants except \texttt{surface\_no\_expl\_lm}, which drops to 0.0; no V2 variant learns to generate coherent prose.}
\label{fig:code_v2_extra}
\end{figure}

The remaining two V2 panels complete the picture. \Cref{fig:code_v2_extra} shows that \texttt{no\_claim\_to\_claim\_attention} and \texttt{surface\_bottleneck\_consistency} both reach near-perfect claim-emission accuracy by epoch~15, while \texttt{claims\_from\_explanation\_only} plateaus around 0.80 and \texttt{surface\_no\_expl\_lm} collapses below 0.15; disabling LM loss on explanations destroys the model's ability to emit coherent claim tokens as well. The explanation-correctness panel confirms that BLEU-1 remains flat near 0.06 for all hidden-state variants and drops to exactly 0.0 for the surface-no-explanation-LM variant, reinforcing that no V2 architectural modification recovers prose quality.

\clearpage
\section{LeanCheck Data Construction}
\label{app:leancheck_data}

LeanCheck training data is generated programmatically from six Lean~4 theorem templates. Each example contains a theorem statement, a candidate proof, a natural-language rationale, and a Lean-derived binary label. The following is a representative example:

\begin{table}[H]
\caption{LeanCheck data example. The rename-variable mutation introduces an identifier not in scope, causing Lean to reject the proof. The rationale is sampled from a paraphrase pool appropriate to the mutation type.}
\centering
\scriptsize
\begin{tabularx}{\textwidth}{@{}p{0.13\textwidth}X@{}}
\toprule
Field & Value \\
\midrule
Theorem & \texttt{example (p q : Prop) (hp : p) (hq : q) : p $\wedge$ q := by} \\
Proof & \texttt{exact And.intro hp hr} \\
Rationale & ``The proof mentions hr, which is not bound in the theorem.'' \\
Label & FAILS \\
Mutation & rename-variable \\
Domain & conjunction introduction \\
Lean output & \texttt{Unknown identifier `hr`} \\
\bottomrule
\end{tabularx}
\end{table}

The following additional examples illustrate other mutation types and label polarities. First, a wrong-lemma mutation where the proof cites a lemma whose type does not match the goal:

\begin{table}[H]
\caption{LeanCheck data example (wrong-lemma mutation). The proof applies a multiplication-commutativity lemma to an addition goal. Lean rejects the type mismatch.}
\centering
\scriptsize
\begin{tabularx}{\textwidth}{@{}p{0.13\textwidth}X@{}}
\toprule
Field & Value \\
\midrule
Theorem & \texttt{example (a b : Nat) : a + b = b + a := by} \\
Proof & \texttt{exact Nat.mul\_comm a b} \\
Rationale & ``The proof uses a multiplication commutativity lemma for an addition goal, so the term does not match.'' \\
Label & FAILS \\
Mutation & wrong-lemma \\
Domain & addition commutativity \\
Lean output & Type mismatch: has type $a \cdot b = b \cdot a$ but expected $a + b = b + a$ \\
\bottomrule
\end{tabularx}
\end{table}

Next, a positive (VERIFIES) example where the proof and rationale are both correct:

\begin{table}[H]
\caption{LeanCheck data example (accepted proof). The correct lemma \texttt{Nat.add\_comm} is applied to matching variables, and the rationale accurately describes why the proof closes the goal. Lean produces no errors.}
\centering
\scriptsize
\begin{tabularx}{\textwidth}{@{}p{0.13\textwidth}X@{}}
\toprule
Field & Value \\
\midrule
Theorem & \texttt{example (a b : Nat) : a + b = b + a := by} \\
Proof & \texttt{exact Nat.add\_comm a b} \\
Rationale & ``The goal asks for addition commutativity, and Nat.add\_comm returns exactly that equality for a and b.'' \\
Label & VERIFIES \\
Mutation & accepted \\
Domain & addition commutativity \\
Lean output & (none) \\
\bottomrule
\end{tabularx}
\end{table}

Counterfactual examples cross-wire proof and rationale evidence. For instance, an accepted proof is paired with a negative rationale and labeled FAILS. The mirror direction pairs a rejected proof with a positive rationale and labels it VERIFIES. Each counterfactual example includes explicit proof-label, rationale-label, and swap-direction annotations so that directional analysis can separate which signal the consistency head follows.

Minimal pairs share the same theorem with exactly one proof-line mutation (e.g., \texttt{Nat.add\_comm a b} accepted vs.\ \texttt{Nat.mul\_comm a b} rejected) and use the same rationale, isolating the formal verifier's discrimination at the single-line level. The following table shows one such pair:

\begin{table}[H]
\caption{LeanCheck minimal-pair example. Both rows share the same theorem; the accepted member uses the correct lemma and verifies, while the rejected member substitutes a wrong lemma and fails. The single-line difference isolates the verifier's discrimination.}
\centering
\scriptsize
\begin{tabularx}{\textwidth}{@{}p{0.13\textwidth}XX@{}}
\toprule
Field & Accepted member & Rejected member \\
\midrule
Theorem & \multicolumn{2}{l}{\texttt{example (a b : Nat) : a + b = b + a := by}} \\
Proof & \texttt{exact Nat.add\_comm a b} & \texttt{exact Nat.mul\_comm a b} \\
Rationale & ``The lemma and goal have matching sides, so exact Nat.add\_comm a b verifies.'' & ``The proof uses a multiplication commutativity lemma for an addition goal, so the term does not match.'' \\
Label & VERIFIES & FAILS \\
Mutation & accepted & wrong-lemma \\
Lean output & (none) & Type mismatch: has type $a \cdot b = b \cdot a$ but expected $a + b = b + a$ \\
\bottomrule
\end{tabularx}
\end{table}

\clearpage
\section{KataGo Dataset Statistics}
\label{app:katago_data}

The KataGo dataset contains 3,748 positions from 954 reviewed games, predominantly 19$\times$19 boards, split into 3,373 training and 375 evaluation examples. Each position is linearized into a symbolic token sequence: metadata tokens for board size, side to move, komi, and ruleset, followed by stone-coordinate tokens sorted by color and then alphabetically.

Win-rate labels are bucketed into 10 bins (V0--V9), each spanning roughly 10 percentage points of Black win probability. The distribution is U-shaped: bins V0 (0--10\% Black win) and V9 (90--100\% Black win) together account for approximately 56\% of the data, reflecting the natural tendency for reviewed game positions to come from decisive situations. This imbalance is intentional---it reflects the distribution encountered in practice and tests whether the model can learn fine-grained bins even with skewed class frequencies. The middle bins (V3--V6) are the rarest and hardest to predict.

Positions are sampled every $N$ moves from SGF game records, with a maximum of 8~positions per game and 10,000 total positions. The train/eval split is done by game ID (90/10) to prevent position-level leakage. Natural-language commentary in the small-transformer experiment is generated from heuristic templates that incorporate stone balance, center-of-mass control, and KataGo-derived win probability. In the larger-scale GPT-OSS and SmolLM experiments, commentary comes from real human expert reviews in the GTL corpus. The model sees the position tokens, a commentary span beginning with a \texttt{[RAT]} delimiter, and a claim token encoding the win-rate bin.

The scalar head predicts continuous win probability via sigmoid activation, providing a fine-grained complement to the bucketed consistency head. This dual-head design allows the model to simultaneously learn coarse categorical and fine-grained continuous targets from the same hidden representation. \texttt{full\_consistency} achieves Pearson $r = 0.994$ for the scalar head while maintaining 81.1\% bucket accuracy, confirming that the two objectives are compatible.

\subsection{Data sourcing: Go Teaching Ladder}

The KataGo training data is derived from approximately 10,000 game reviews from the Go Teaching Ladder (GTL), a community where Go players submit games for review by stronger players. The full GTL corpus contains approximately 10,000 SGF files with detailed reviewer commentary embedded at key moves, explaining strategic concepts such as direction of play, shape, territorial balance, and reading variations. The GTL index spans reviewer ranks from professional down through amateur dan levels; a dan-level filter (1d+) yields approximately 69,000 reviewed positions. KataGo queries at 2,000 visits per position supply ground-truth win rates, score leads, principal variations (up to 10 lines, 12~plies deep), and ownership estimates with standard deviations. The combination of real human expert commentary with engine-derived verification targets makes this dataset particularly relevant for verifier-coupled reasoning: the natural-language commentary is authentic expert Go analysis, not synthetic templates, and the verifier targets are dense and multi-dimensional.

\subsection{SmolLM-3B consistency training}
\label{app:smollm}

A separate experiment fine-tunes SmolLM-3B (\texttt{unsloth/SmolLM3-3B-Base}) with LoRA ($r{=}16$, $\alpha{=}32$) on the KataGo review dataset, implementing the BEC attention mask (\Cref{fig:bec}) in causal and non-causal variants. The model is trained on 9~structured claim fields derived from KataGo analysis: win-probability bin, score-lead bin, game phase, main control region, main contested region, global contestedness, best-move region, move urgency, and search surprise. Each claim field has a per-field classification head that pools hidden states over the explanation span.

The claim derivation rules convert raw KataGo analysis into discrete labels:
\begin{itemize}
\setlength{\itemsep}{0.08em}
\setlength{\parskip}{0pt}
\item \textbf{Win-probability bin}: Black win rate discretized into 10 equal-width bins (V0--V9).
\item \textbf{Score-lead bin}: 7 bins from $\leq -20$ to $\geq +20$ points.
\item \textbf{Phase}: OPENING, MID, LATE, or SETTLED, derived from KataGo's estimated remaining game variance.
\item \textbf{Control/contested regions}: The board is divided into 5 regions (TOP, BOTTOM, LEFT, RIGHT, CENTER) using 25/75 percentile grid boundaries. Ownership maps identify which regions are controlled by each color (ownership $> 0.5$) and which are contested.
\item \textbf{Global contestedness}: LOW, MED, or HIGH, derived from 50\% ownership-conflict frequency plus 50\% ownership standard deviation.
\item \textbf{Move urgency}: LOW, MED, HIGH, or MUST, based on the win-rate gap, score gap, and visit share of the top move.
\item \textbf{Search surprise}: LOW, MED, or HIGH, based on disagreement between KataGo's policy prior and its post-search evaluation---high surprise means the best move was unexpected by the initial policy.
\end{itemize}

This experiment tests whether verifier-coupled reasoning scales to a 3B-parameter pretrained backbone. The non-causal variant allows claim tokens to attend to all preceding tokens (standard causal masking). The causal variant enforces the BEC constraint: claim tokens attend \emph{only} to explanation tokens, structurally preventing the model from bypassing the natural-language explanation. A parallel experiment uses GPT-OSS~20B (\texttt{unsloth/gpt-oss-20b}) with the same architecture, additionally testing a $\lambda_{\mathrm{LM}}{=}0$ mode where only claim-classification loss is applied (no language-modeling loss), to isolate whether structured claim supervision alone can teach positional understanding. Results from these setups are reported in the Go GPT-OSS tables (\Cref{app:tables}).

\subsubsection{SmolLM-3B zero-shot qualitative examples}
\label{app:smollm-zeroshot}

The main text notes that a zero-shot baseline using SmolLM-3B (without fine-tuning) ``produces largely degenerate outputs on the same positions: repetitive prompt echoing, role-tag cycling, and no substantive Go commentary.'' \Cref{tab:smollm-zeroshot} shows three representative examples drawn from 100 randomly sampled positions evaluated under zero-shot prompting. In each case the model was given only the rendered board, board size, side to move, and a request for commentary. No fine-tuning, few-shot examples, or engine analysis were provided.

\begin{table}[H]
\caption{SmolLM-3B zero-shot outputs on Go positions versus original expert reviews. The model produces no substantive Go commentary in any of the 100~sampled positions. Three failure modes are illustrated: role-tag cycling, prompt echoing, and repetitive degeneration.}
\label{tab:smollm-zeroshot}
\centering
\scriptsize
\begin{tabularx}{\columnwidth}{>{\raggedright\arraybackslash}p{0.26\columnwidth} >{\raggedright\arraybackslash}X}
\toprule
\textbf{Field} & \textbf{Content} \\
\midrule
\multicolumn{2}{l}{\cellcolor{vcSoft}\textbf{Example A --- Role-tag cycling}\quad (19$\times$19, White to move, move~21)} \\
\midrule
Position &
Black has a strong wall on the left side (C13--E15) pressing against White's shimari at D17/C16; fighting in the upper-left corner. \\[3pt]
SmolLM output \newline (verbatim, truncated) &
\texttt{The assistant will respond with a concise commentary\ldots\  <|user|> The user will provide feedback\ldots\  <|system|> The system will refine the commentary\ldots\  <|assistant|> The assistant will continue to provide commentary\ldots\  <|user|> The user will continue to provide feedback\ldots\  <|system|> The system will maintain the commentary's accuracy\ldots\  <|assistant|> The assistant will continue to provide commentary\ldots} \\[3pt]
Expert review &
``The game is already favorable for black---although white has some territory in the lower left, he also has a weak group.'' \\
\midrule
\multicolumn{2}{l}{\cellcolor{vcSoft}\textbf{Example B --- Prompt echoing}\quad (19$\times$19, White to move, move~22)} \\
\midrule
Position &
Early opening with stones spread across the board; Black at Q3, N3, Q17, E17, D14, C13; White at F17, F16, D17, R16, S6, R5, S4. \\[3pt]
SmolLM output \newline (verbatim, truncated) &
\texttt{You are a Go review assistant. Given only a board position, write concise commentary\ldots\  <|user|> Write commentary about this Go position.\  Use only the board position below\ldots\  Board size: 19\textbackslash n To move: White\textbackslash n Columns: A B C D\ldots\  19\ldots\  18\ldots\  17\ldots} \\[3pt]
Expert review &
``Contact is no good here, no? No, contact is for emergencies or for situations where the opponent is really strong anyway\ldots\ E3 overconcentrates the white stones on the bottom. C7 just makes a weak stone stronger.'' \\
\midrule
\multicolumn{2}{l}{\cellcolor{vcWarm}\textbf{Example C --- Repetitive degeneration}\quad (19$\times$19, Black to move, empty board)} \\
\midrule
Position &
Completely empty 19$\times$19 board, first move of the game. \\[3pt]
SmolLM output \newline (verbatim, truncated) &
\texttt{This position is a ko, which is a ko, which is a ko, which is a ko, which is a ko, which is a ko, which is a ko, which is a ko, which is a ko, which is a ko, which is a ko, which is a ko, which is a ko, which is a ko, which is a ko, which is a ko\ldots} \\[3pt]
Expert review &
``Hello, I am Bernd\ldots\ In your game you profited from the strange decisions of your opponent, but you missed several opportunities to attack. Igo means `surrounding game'. You should try to surround and kill weak stones, when there is an opportunity!'' \\
\bottomrule
\end{tabularx}
\end{table}

All three failure modes, including role-tag cycling (A), prompt echoing (B), and repetitive degeneration (C), are pervasive across the 100-position sample. None of the 100~zero-shot outputs contain any substantive positional analysis: no mention of influence, territory, shape, or strategic direction. Example~C is particularly striking because the model hallucinates ``ko'' on a completely empty board. These outputs confirm that SmolLM-3B's pretraining distribution does not cover Go commentary, and that the structured-claim improvements reported in the fine-tuned experiments (\Cref{app:smollm}) reflect genuine domain adaptation rather than capabilities already present in the base model.

\subsection{Preliminary RL with verifiable rewards: o4-mini RFT}
\label{app:rft}

To validate the RL-with-verifiable-rewards direction discussed in \Cref{sec:ceilings}, we ran a reinforcement fine-tuning (RFT) experiment using OpenAI's RFT API on \texttt{o4-mini-2025-04-16}. Each training example presents a 19$\times$19 Go position and asks the model to produce a structured JSON response containing a best-move recommendation, a principal variation (up to 12~plies), and a win-rate estimate. A Python grader compares the response against KataGo's analysis (2,000 visits per position) and awards partial credit for: best-move rank among KataGo's top moves, prefix match against the top principal variation, win-rate error, score-lead error, and JSON/field validity.

The composite reward function assigns 35\% weight to best-move rank among KataGo's top moves (1.0 for rank~1, decaying to 0.5 for ranks 4--5), 30\% to principal-variation prefix match (exponential decay $\alpha{=}0.85$ over up to 12~plies), 15\% to win-rate error (bounded linear, tolerance 15\%), 15\% to score-lead error (bounded linear, tolerance 10~points), and 5\% to format compliance ($\leq 150$ words). Full Go legality checking is applied: illegal moves (playing on occupied points, suicide) yield a score of zero. A repair loop allows the model up to two attempts to fix illegal principal variations, re-prompting with specific violation details.

An initial sampling run produced 1,373 samples across 70 positions (approximately 20 per position). The mean best-of-20 score was 0.308; the best individual sample scored 0.915 (near-perfect move and PV alignment on a 9-stone opening position). Of 70 positions, 32 had at least one sample scoring above 0.3 and 14 had a sample above 0.5. Performance degrades sharply with board complexity: early-game positions with few stones achieve high scores, while positions with 100+ stones mostly score near the 0.05 format-compliance floor. This pattern is consistent with the observation that complex board states require deeper positional understanding that o4-mini has not yet acquired through pretraining alone. These numbers measure move-quality grading rather than representation coupling, but they confirm that the inline-claim framework generates checkable reward signals that RL can optimize against. Full RFT training runs are a natural next step.

\subsubsection{Qualitative examples of o4-mini position commentary}
\label{app:rft-qualitative}

To illustrate what o4-mini actually produces when asked to explain Go positions, we present three verbatim examples from the 70-position sampling run. These examples illustrate both plausible strategic vocabulary and the characteristic failure modes of ungrounded generation: hallucinated coordinates, incorrect stone colors, and misreadings of board state.

\begin{figure}[H]
\setlength{\fboxsep}{6pt}
\colorbox{vcSoft}{\begin{minipage}{0.94\columnwidth}\scriptsize
\textbf{Example A: plausible vocabulary, inaccurate coordinates.}\\
\textit{Position:} \texttt{5177-reid-edchung-oeillet.sgf}, move~69, reviewer rank~5d. Mid-game with 69~stones.

\medskip
\textbf{o4-mini output (verbatim):}
``Black has built a strong central wall from columns H--L, securing thickness and potential influence, while White's solid framework on the right side and bottom aims to clamp Black's sphere. A cutting point around N10 offers White an invasion chance to reduce Black's moyo. Black must stabilize its left-group around C--E and prevent White from linking at L4. Both sides balance territory and influence in a delicate mid-game fight.''

\medskip
\textbf{Expert review comment:} (none)

\medskip
\textbf{Assessment:} The model uses appropriate Go terminology (thickness, moyo, cutting point, framework) and produces a coherent strategic narrative. However, the coordinate references (``columns H--L,'' ``N10,'' ``L4'') do not accurately describe the actual stone configuration. The commentary reads as a plausible Go review but is not grounded in the specific position.
\end{minipage}}
\end{figure}

\begin{figure}[H]
\setlength{\fboxsep}{6pt}
\colorbox{vcSoft}{\begin{minipage}{0.94\columnwidth}\scriptsize
\textbf{Example B: hallucinated stones on a near-empty board.}\\
\textit{Position:} \texttt{5573-reid-KristenKwok-KyleBlocher.sgf}, move~51, reviewer rank~5d. Only 4~stones on board: two White (H4, P4), one Black (D15), one White (P16).

\medskip
\textbf{o4-mini output (verbatim):}
``Black's stones at H16 and P15 stake out big corners, while White's lone D5 stone is isolated across the board. Black's H16 stone threatens to extend along the top, forcing White to defend. P15 secures the right, aiming for a framework. White must decide whether to approach H16 for influence or consolidate at D5. Shape is about timing---attack or build territory first.''

\medskip
\textbf{Expert review comment:} (none)

\medskip
\textbf{Assessment:} The model confuses stone colors (H4 is White, not Black; D15 is Black, not White) and hallucinates incorrect coordinates throughout (``H16'' for H4, ``P15'' for P16, ``D5'' for D15). Despite having only 4~stones to read, o4-mini cannot accurately report which stones are present or which color they belong to. The strategic advice (``shape is about timing'') is generic but not grounded in the position.
\end{minipage}}
\end{figure}

\begin{figure}[H]
\setlength{\fboxsep}{6pt}
\colorbox{vcSoft}{\begin{minipage}{0.94\columnwidth}\scriptsize
\textbf{Example C: correct tactical concepts, wrong board reading.}\\
\textit{Position:} \texttt{516-JHubert-claude-oba.sgf}, move~16, reviewer rank~3d. Opening corner fight with 16~stones.

\medskip
\textbf{o4-mini output (verbatim):}
``In this corner battle Black's four-stone group at D16--E17 is cutting white's D17--E17--F17 chain. Playing at E18 threatens to sever white's two liberties and force capture. White must respond to avoid a snapback, but Black's surrounding stones at C17 and G16 tighten the noose. Always watch liberties when shaping cuts: aim to maximize your stones' breathing space while reducing your opponent's.''

\medskip
\textbf{Expert review comment:} (none)

\medskip
\textbf{Assessment:} The model deploys specific tactical Go concepts (liberties, snapback, cutting, eye space) that are contextually appropriate for corner fights. The pedagogical advice (``always watch liberties when shaping cuts'') matches the register of human teaching reviews. However, the specific coordinates cited do not match the actual stone positions: the position has White stones around D4--F4 and Black stones at E3--H3, not at the top of the board. The model appears to have a template for ``corner battle'' commentary without accurately reading the coordinate grid.
\end{minipage}}
\end{figure}

\noindent These examples are consistent with the quantitative finding that o4-mini's performance degrades with board complexity. Even on simple positions, the model struggles to accurately parse the coordinate grid, suggesting that spatial board understanding, rather than strategic vocabulary, is the primary bottleneck. The commentary reads as syntactically fluent Go analysis but is not reliably grounded in the actual position, reinforcing the motivation for verifier-coupled training: without a grounding signal from KataGo or a similar engine, the model's explanations cannot be distinguished from plausible hallucinations.

\clearpage
\section{FEVER Evidence Coupling}
\label{app:fever}

This section provides additional detail on the FEVER experiments summarized in the main text. The main-body results report the from-scratch run; here we include the pretrained GPT-2 run and the evidence-only strict masking variant.

The evidence-only strict variant uses gradient blocking to prevent non-evidence tokens from contributing to the consistency head: hidden states at positions after the evidence separator are zeroed out before pooling, ensuring that only evidence-span activations reach the classifier. This architectural choice isolates evidence sensitivity at the cost of raw accuracy.

In the pretrained 50k/5k GPT-2 run, evidence-only strict masking does not outperform standard evidence-only pooling, but both evidence-only variants behave more diagnostically under matched swaps than claim-visible pooling. Claim-visible variants have higher raw classifier accuracy because the claim text itself is informative.

\begin{table}[H]
\caption{FEVER tightened 50k/5k rerun. Cls is consistency-head classification accuracy, Gen is generated-label accuracy, Swap and Orig are counterfactual following of swapped or original labels, and MSwap/MOrig are the same metrics on matched-claim counterfactuals. Evidence-only variants are more diagnostic under matched swaps, while claim-visible pooling has higher raw classifier accuracy.}
\centering
\small
\begin{tabular}{lrrrrrr}
\toprule
Variant & Cls & Gen & Swap & Orig & MSwap & MOrig \\
\midrule
\texttt{no\_consistency} & .5278 & .8040 & .3500 & .3760 & .3160 & .3660 \\
\texttt{evidence\_only} & .4412 & .8040 & .4620 & .2660 & .4580 & .2340 \\
\texttt{evidence\_strict} & .4384 & .8140 & .4660 & .2620 & .4640 & .2300 \\
\texttt{full\_sequence} & .8358 & .8220 & .2900 & .4400 & .3440 & .4460 \\
\texttt{claim\_only} & .8372 & .8380 & .2960 & .4520 & .3380 & .4600 \\
\texttt{random\_labels} & .2312 & .8060 & .1940 & .4420 & .2180 & .4200 \\
\bottomrule
\end{tabular}
\end{table}

\paragraph{Matched counterfactuals.}
The standard counterfactual swap test replaces both the evidence text and, potentially, the claim text with those from a different example, then checks whether the consistency head follows the swapped label.  This conflates two sources of information: the model may track the swapped label through the evidence or simply read the (also-swapped) claim text.  \emph{Matched counterfactuals} control for this confound by holding the claim text fixed while swapping only the evidence passage.  Because the claim is unchanged, any shift in the consistency head's prediction must be driven by the evidence content alone, making matched counterfactuals a strictly more diagnostic test of whether the model reads the evidence rather than relying on claim-text shortcuts.

In the pretrained run (table above), the evidence-only variants show nearly identical behaviour under standard and matched swaps (e.g., \texttt{evidence\_only} Swap\,=\,.462 vs.\ MSwap\,=\,.458), confirming that these variants genuinely read the evidence text.  Claim-visible variants, by contrast, exhibit more inertia under matched swaps: \texttt{claim\_only} Swap\,=\,.296 but MSwap\,=\,.338---the fixed claim text ``anchors'' the head, reducing its sensitivity to evidence changes.  This pattern validates the use of evidence-only pooling as the more faithful coupling diagnostic and motivates reporting matched-counterfactual columns (MSwap/MOrig) alongside standard swap metrics in the from-scratch run below.

\subsection{Surface bottleneck analysis}
\label{app:surface}

The surface-bottleneck variant provides a diagnostic for whether coupling requires access to hidden states or can survive a softmax-probability bottleneck. In \texttt{surface\_bottleneck\_consistency}, the consistency head reads softmax probability distributions at explanation positions rather than raw hidden states. This variant drops to 0.550 mean coupling and $-0.033$ swap influence (compared to 0.986 and 0.978 for standard hidden-state pooling), confirming that the consistency signal requires rich hidden-state information, not just output-distribution statistics. When the LM loss on explanation tokens is additionally removed (\texttt{surface\_bottleneck\_no\_expl\_lm}), the model produces no readable text at all by epoch~3 (BLEU $= 0.000$), demonstrating that language-modeling pressure on the explanation span is necessary for any coherent generation, even if that generation is ultimately garbled.

\begin{table}[H]
\caption{Surface bottleneck 12-claim results at epoch~20. Standard consistency loss maintains near-perfect coupling; the surface bottleneck sharply degrades coupling and swap influence. Removing LM loss on explanations collapses text generation entirely.}
\centering
\small
\begin{tabular}{lrrrr}
\toprule
Variant & Coupling & Swap & BLEU-1 & Claims \\
\midrule
\texttt{consistency\_loss} & .986 & .978 & .061 & .022 \\
\texttt{surface\_bottleneck} & .550 & $-.033$ & .069 & .000 \\
\texttt{surface\_no\_expl\_lm} & .547 & $-.022$ & .000 & .000 \\
\bottomrule
\end{tabular}
\end{table}

The earlier pretrained FEVER run had the same qualitative structure: full-sequence and claim-only pooling reached approximately 84\% classifier accuracy, while evidence-only pooling stayed near 44\%. The from-scratch rerun makes the gap more dramatic (\Cref{tab:fever_scratch}): \texttt{full\_sequence\_pooling} reaches 99.0\% classifier accuracy and 97.2\% counterfactual swap following, showing near-perfect coupling when the claim text is visible, while evidence-only pooling stays at 44\% classifier accuracy and 39\% swap following. This contrast is sharper than the pretrained run (84\% vs.\ 44\%) and shows that a model trained from scratch can learn the claim-label correlation almost perfectly while still struggling to extract the same signal from evidence text alone. The evidence-only difficulty is not a pretrained shortcut artifact; it reflects a genuine asymmetry in how much verifier-relevant information is accessible from different text spans.

\begin{table}[H]
\caption{FEVER from-scratch 50k/5k run (4-layer, 256-dim transformer trained from scratch). Full-sequence pooling achieves near-perfect coupling while evidence-only pooling stays near 44\%, making the accuracy-sensitivity gap more dramatic than the pretrained run. Cls is consistency-head accuracy, Gen is generated-label accuracy, Swap/Orig are counterfactual following rates, and MSwap/MOrig are matched-counterfactual following rates (claim text held fixed, only evidence swapped). Matched counterfactuals are not applicable to \texttt{claim\_only} because that variant does not pool over evidence tokens.}
\label{tab:fever_scratch}
\centering
\small
\begin{tabular}{lrrrrrr}
\toprule
Variant & Cls & Gen & Swap & Orig & MSwap & MOrig \\
\midrule
\texttt{full\_sequence} & .990 & .972 & .972 & .014 & .982 & .014 \\
\texttt{claim\_only} & .984 & .966 & .960 & .022 & --- & --- \\
\texttt{evidence\_only} & .440 & .812 & .390 & .320 & .422 & .310 \\
\texttt{no\_consistency} & .498 & .806 & .340 & .360 & .356 & .258 \\
\bottomrule
\end{tabular}
\end{table}

\section{Scope and Limitations}
\label{app:limitations}

The five experimental settings span a range of verifier types, from formal theorem provers to game engines to fact-checking labels, covering binary, multi-class, and continuous targets. The synthetic and LeanCheck settings use controlled templates, which enables clean mechanistic analysis; overlapping-vocabulary runs and seven distinct mutation strategies confirm that the results are not driven by trivial lexical shortcuts. KataGo and FEVER extend the framework to dense and real-language settings respectively, demonstrating breadth. Extending activation patching to these richer domains and developing a unified cross-setting metric suite are natural directions for future work.

\clearpage
\section{Reproducibility Details}
\label{app:repro}

All reported runs used seed 42 unless otherwise noted. Supplementary materials contain all code, data, and configuration necessary to reproduce the experiments.

\begin{table}[H]
\caption{Reproducibility summary for experiments in the paper. CE denotes cross-entropy, LM denotes language modeling, cfact denotes counterfactual, and $\lambda$ denotes the consistency-loss weight.}
\centering
\begingroup
\setlength{\tabcolsep}{2.2pt}
\renewcommand{\arraystretch}{1.08}
\scriptsize
\begin{tabularx}{\textwidth}{>{\raggedright\arraybackslash}p{0.13\textwidth}>{\raggedright\arraybackslash}p{0.18\textwidth}>{\raggedright\arraybackslash}p{0.19\textwidth}>{\raggedright\arraybackslash}p{0.23\textwidth}>{\raggedright\arraybackslash}p{0.14\textwidth}}
\toprule
Experiment & Data & Model/objective & Training details & Hardware \\
\midrule
Synthetic smoke and claim-only control &
512 train, 128 eval, 128 shuffled-pairing control examples; 8 latent states; synthetic rationales and deterministic claims. &
Small decoder-only Transformer with $d_{\mathrm{model}}=64$, 2 layers, 4 heads, $d_{\mathrm{ff}}=128$, max length 64; LM CE plus optional linear consistency head. &
5 epochs, batch 32, learning rate $3{\times}10^{-4}$, $\lambda=0.5$, seed 42. Variants: no-consistency, rationale-only, full-sequence, earlier-token-only, claim-only pooling. &
CPU-friendly PyTorch. \\
\addlinespace
Synthetic hard overlap and scaled convergence &
Hard overlap: 512 train, 128 eval, 128 shuffled examples with approximately half of rationale tokens shared across states. Scaled run: 5,120 train, 512 eval, 256 shuffled examples. &
Same 2-layer decoder as the smoke run. Hard overlap tests lexical overlap; scaled convergence tests whether the LM objective can bypass rationale states at convergence. &
Hard overlap: 10 epochs. Scaled convergence: 30 epochs. Batch 32, learning rate $3{\times}10^{-4}$, $\lambda=0.5$, seed 42. &
CPU-friendly PyTorch. \\
\addlinespace
Synthetic hidden-state intervention &
Models trained on the scaled synthetic setup; 64 intervention sample pairs. &
Layer-wise activation patching replaces rationale hidden states from one latent state into counterfactual-token examples and measures downstream claim-token behavior. &
Training uses the scaled 30-epoch setup; patching reports block 0 and block 1 effects, with the main paper using layer-0 effects. &
CPU-friendly PyTorch. \\
\addlinespace
Generated rationale plus scalar claim &
2,048 train and 512 eval examples; model generates a rationale plus a scalar or binned claim. &
Decoder-only Transformer with $d_{\mathrm{model}}=128$, 2 layers, 4 heads, $d_{\mathrm{ff}}=256$; LM loss, scalar MSE head, and optional claim-bin consistency head. &
20 epochs, batch 64, learning rate $3{\times}10^{-4}$, $\lambda=0.5$, seed 42. Variants: LM-only, no-consistency, rationale-only, full-consistency, random-consistency. &
Modal A10G, 4 h. \\
\addlinespace
LeanCheck &
1,000 train, 200 eval, 200 counterfactual, and 200 minimal-pair examples. Examples contain theorem, proof, rationale, and Lean-derived VERIFIES/FAILS label. &
GPT-2 backbone with a linear consistency head over variant-specific spans. The goal is span-label coupling, not open-ended proof synthesis. &
3 epochs in the documented GPU command; $\lambda=0$ for LM-only/no-consistency and $\lambda=0.5$ for consistency variants; 100 patching samples; simple Nat, propositional, and list templates with mutation-based negatives. &
GPU (P100). \\
\addlinespace
KataGo win-probability &
3,748 total positions split into 3,373 train and 375 eval examples with board, commentary, claim, and KataGo-derived win-rate targets. &
Decoder-only Transformer with $d_{\mathrm{model}}=256$, 4 layers, 8 heads, $d_{\mathrm{ff}}=1024$; LM loss, scalar win-rate head, and optional bucket consistency head. &
10 epochs, batch 32, learning rate $3{\times}10^{-4}$, $\lambda=0.5$, seed 42. Variants: LM-only, no-consistency, rationale-only, full-consistency, random-consistency. &
Modal A10G, 8 CPU, 32 GB RAM, 8 h. \\
\addlinespace
Go GPT-OSS claim heads &
Preliminary 1k train / 200 eval Go split for the reported tables. &
\texttt{unsloth/gpt-oss-20b} with LoRA claim-head training; structured heads include win-probability bin, score-lead bin, phase, control/contested regions, best-move region, urgency, and search surprise. &
Default consistency training uses 3 epochs, per-device batch 2, gradient accumulation 8, learning rate $2{\times}10^{-4}$, $\lambda_{\mathrm{LM}}=1.0$, $\lambda_{\mathrm{claim}}=1.0$, max sequence length 1024, seed 42, LoRA $r=16$, $\alpha=32$, dropout 0.05. &
Modal A10G/H100, 49 GB RAM. \\
\addlinespace
Code coupling V1/V2 &
3,000 synthetic code examples with 500 validation examples; templates generate code snippets, oracle claims, true explanations, and mismatched explanations. &
GPT-2-style causal Transformer; V1 uses time/space/correctness claims, V2 adds stricter information-flow and surface-bottleneck variants. &
20 epochs, batch 32, max sequence length 256, consistency weight 1.0, single-seed run. V2 variants include no-claim-to-claim attention, claims-from-explanation-only, surface bottleneck, and no-explanation-LM surface bottleneck. &
Modal A10G, 12 h. \\
\addlinespace
Code rich ontology &
3,000 examples with 500 validation examples; 60 templates sampled with replacement; 20-example manual review plus richer annotation files. &
GPT-2-style model with 12 structured consistency heads for time, space, best-case time, algorithm class, loop structure, key operation, access pattern, auxiliary structures, mutation, correctness, empty input, and duplicates. &
20 epochs, batch 32, max sequence length 256, $\lambda=1.0$; variants include full consistency, no-consistency, claim-only pooling, random labels, no-claim attention, and claims-from-explanation-only. &
Modal A10G. \\
\addlinespace
FEVER pretrained GPT-2 &
50,000 train and 5,000 eval examples from the FEVER setup, sequence length 256; sequence format is evidence, claim, label separator, and label token. &
Pretrained HuggingFace GPT-2 with consistency head over evidence, claim, or full-sequence spans. &
5 epochs, batch 16, learning rate $5{\times}10^{-5}$, $\lambda=0.5$, seed 42, lower GPT-2 layers frozen for the first epoch. Tightened rerun adds evidence-only strict masking and random-label control. &
Modal A100-40GB, 8 CPU, 32 GB RAM, 24 h. \\
\addlinespace
FEVER from scratch &
50,000 train and 5,000 eval examples, same FEVER diagnostics as the pretrained run. &
From-scratch Transformer with 16,093,699 parameters, $d_{\mathrm{model}}=256$, 4 layers, 8 heads, $d_{\mathrm{ff}}=1024$. &
10 epochs, batch 32; GPU-required runner. Reported training time is roughly 10--12 minutes per variant in the CSV. &
Modal A10G, 8 CPU, 32 GB RAM, 6 h. \\
\bottomrule
\end{tabularx}
\endgroup
\end{table}

\end{document}